\documentclass[12pt]{article}
 \usepackage{algorithmic}
 \usepackage{algorithm}
 \usepackage{amsmath,amssymb,amsthm,mathrsfs,amsfonts}
 \usepackage{bm}
 \usepackage{latexsym}
 \usepackage{CJK}
 \usepackage{mathtools}
 \usepackage{multirow}
 \usepackage{multicol}
 \usepackage{caption}
\usepackage{graphicx,color}
\usepackage{amssymb,amsfonts}
\usepackage{hyperref}
\usepackage[english]{babel}
\usepackage{times}
\usepackage[T1]{fontenc}
\usepackage{enumerate}
\usepackage{geometry}
\usepackage{float}
\usepackage{appendix}
\usepackage{booktabs}
\usepackage{endnotes}
\usepackage{setspace}
\usepackage{stmaryrd}
\usepackage[english]{babel}
\usepackage[sectionbib]{natbib}

\newtheorem{thm}{Theorem}
\newtheorem{lemma}{Lemma}
\newtheorem{prop}{Proposition}

\def\argmin{\arg\min}

\begin{document}
\title{Individualized Multilayer Tensor Learning with An Application in Imaging Analysis}
\medskip
\author{Xiwei Tang $^{1}$ , Xuan Bi $^{2}$  and Annie Qu $^{3}$ \footnotemark[1]
}
\bigskip

\date{}\maketitle

\renewcommand{\thefootnote}{\fnsymbol{footnote}}

\footnotetext[1] { $^{1}$ Xiwei Tang is Assistant Professor, Department of Statistics, University of Virginia,
VA 22903 (E-mail: xt4yj@virginia.edu). $^{2}$ Xuan Bi is Assistant Professor, Information and Decision Sciences, Carlson School of Management, University of Minnesota, MN 55455 (Email:xbi@umn.edu). $^{3}$ Annie Qu is Professor, Department of Statistics, University
of Illinois at Urbana-Champaign, Champaign, IL 61820 (E-mail: anniequ@illinois.edu).}

\begin{abstract}
\singlespacing

This work  is motivated by multimodality breast cancer imaging data, which is quite challenging  in that the signals of discrete  tumor-associated microvesicles (TMVs) are randomly distributed  with  heterogeneous patterns.  This  imposes a significant challenge  for conventional imaging regression and dimension reduction models assuming a homogeneous feature structure.
We develop an innovative multilayer tensor learning method to  incorporate heterogeneity to a higher-order tensor decomposition and predict disease status effectively through utilizing subject-wise imaging features and multimodality information.  Specifically, we construct a multilayer decomposition which leverages an individualized imaging layer in addition to a modality-specific  tensor structure.
One major advantage of our approach is that we are able to efficiently  capture the heterogeneous  spatial features of signals that are not characterized by a population structure as well as integrating  multimodality  information simultaneously.
To achieve scalable computing, we develop a new bi-level block improvement algorithm. In theory, we investigate both the algorithm convergence property, tensor signal recovery error bound and   asymptotic consistency for prediction model estimation. We also apply the proposed method for  simulated  and human breast cancer imaging data. Numerical results demonstrate that the proposed method outperforms other existing competing methods.

\bigskip


\end{abstract}

\newpage

\section{Introduction}

In recent years imaging analysis has encountered  explosive growth due to high demand and applications in biomedical imaging for diagnosing disease status and assessing treatment outcomes. The biomedical applications of imaging analyses are especially powerful in cancer radiotherapy and neuroimaging  \citep{Fass:2008, Martino:2008, Caffo:2010, Zhou:2013, Li:2015}. In addition, new advanced technologies  in imaging analysis can be utilized to better understand the  structures of the human body associated with biological, psychological and clinical traits, \citep{Lindquist:2008, Lazar:2008, Friston:2009, Hinrichs:2009, Kang:2012},  so cancer and  chronic diseases can be diagnosed earlier and  intervention  treatments  can be implemented.

\begin{figure}[H]
     \centering
     \includegraphics[height=8cm]{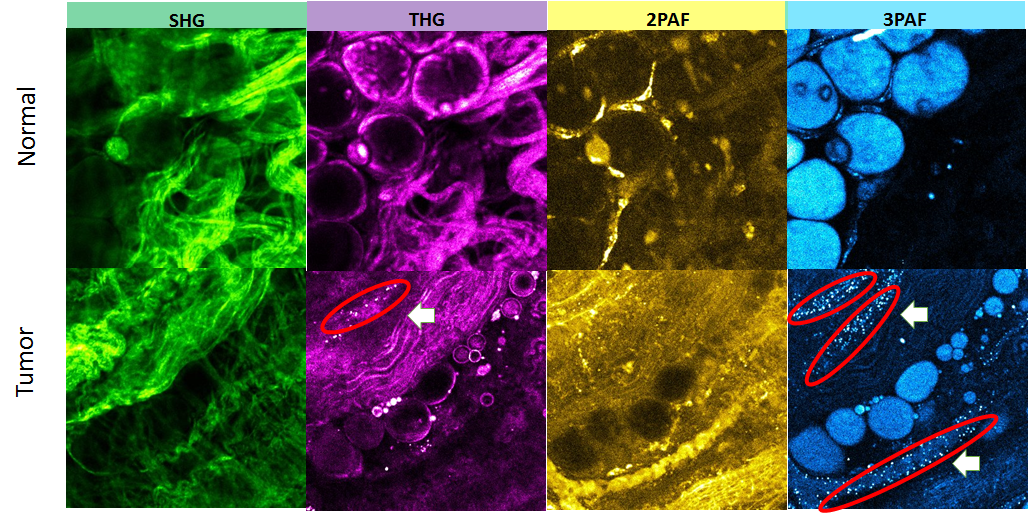}
     \caption{The four-modality microscope images for a normal rat's tissue and a cancerous rat's tissue. The increased amount and appearance of TMVs are clearly evident in the cancerous breast  tissue, compared to the normal breast tissue (see red circles of TMVs).}
     \label{fig:rat}
\end{figure}

This paper is motivated  by  utilizing  multimodality optical imaging data for diagnosing  early-stage breast cancer  prior to  tumor formation.  Figure \ref{fig:rat} illustrates  four-modality optical imaging data for cancerous  and  normal rat breast tissues,  showing  that a large number of tumor-associated microvesicles (TMVs; circled in red) appear spatially aligned in  the cancerous tissue.
One unique aspect of cancer imaging is that individuals might have imaging at various locations, and TMVs as an important biomarker are also randomly located leading to   heterogeneous signal patterns.   This is quite different from the brain imaging where the same regions of the brain usually share similar characteristics and functions over different subjects.  Moreover,  TMVs have relatively weak signals in terms of  smaller size and pixel contrast values compared to modality-specific background and noise.  Both great heterogeneity and  weak strength  make the signals  difficult to be captured by a conventional  population model  which mainly characterizes the population-wise variation.  In addition, it is also crucial to integrate information from multiple imaging  modalities effectively.

In contrast to standard vector data, imaging data are usually in the form of a multi-dimensional array (also known as a multi-order tensor),  which preserves  higher-order structures containing  rich spatial information.   Traditional statistical  models mostly  treat  covariates as vectors.  However, fitting a model with the vectorized  imaging  array  becomes infeasible
when the array size is large.  For instance,  a four-modality  breast cancer imaging size  of $3,600 \times 3,600$ for each modality implicitly requires $4\times3,600^2 = 51,840,000$ regression parameters. This leads to an ultra-high dimensional problem which could be unestimable even with additional regularization techniques. Most importantly, vectorization of an array is not capable of preserving the important spatial structure of imaging.  Under the vectorization framework, Bayesian variable selection approaches have been  developed for  high-dimensional imaging  regression models to identify  important regions, by applying  Markov random field priors to account for the spatial correlation between voxels \citep{Penny:2005, Bowman:2007, Bowman:2008, Derado:2010, Li:2015}.  However, this would rapidly increase the complexity of the prior  when the tensor order increases, and could be computationally infeasible if the tensor order is high.

Alternatively, functional data analysis can be adopted to construct a two-dimensional image predictor \citep{Reiss:2010} in a functional regression model. Nevertheless, the extension to three dimensions or beyond could be impractical due to high-dimensional parameters arising from  higher-order imaging data \citep{Zhou:2013}.  Recent  developments in imaging analyses employ a two-stage strategy, that is, perform a dimension reduction such as principle component analysis (PCA) first, and then fit a regression model based on the extracted principal components \citep{Caffo:2010},  however, which still  requires conversion of the image into a vector first.

To utilize the unique higher-order  structure of the image covariates, \cite{Zhou:2013} and \cite{Li:2013} propose a tensor regression model, where the coefficients associated with imaging  voxels are formulated as  a tensor and assumed  to  be   low-rank.   But this  potentially requires  voxel-wise registration over different individuals, which is not  applicable for cancer imaging.  In addition, the tensor decomposition technique,  such as the  CANDECOMP/PARAFAC (CP)  method \citep{Bechmann:2005},   is applied for processing  signals from the image covariates  directly  \citep{Cong:2012, Morup:2006, Karahan:2015, Lee:2007, Miranda:2015}.  Due to its population-wise low-rank constraint, it is not capable of  capturing heterogeneous imaging information.

 Another prevalent tool in image recognition and classification  is the deep learning method,  in particular the convolutional neural network  \citep[CNN,][]{Ciresan:2011, Krizhevsky:2013}.  In general, the CNN is  powerful for image classification, which utilizes multiple hidden layers with a variety  of convolution, pooling and activation methods.  However,   due to its complex architecture and the large number of parameters involved, the CNN usually requires massive training data to guarantee  good  performance \citep{Keshari:2018, Wagner:2013, Yu:2017b, Yu:2017a},  especially when the imaging signals are heterogeneous and relatively weak.   In our motivated cancer imaging study, such a large sample size for homogeneous samples is infeasible.

In this paper, we propose  an individualized multilayer  tensor learning  (IMTL) utilizing additional individualized layers to incorporate heterogeneity to a higher-order tensor decomposition, which captures extra variation of the imaging covariates that cannot be characterized  by a population structure.
Furthermore,   we extend it to multimodality data to integrate individual-specific information over different modalities and  incorporate modality-specific features through different layers.   We develop a scalable and efficient algorithm which can be implemented with parallel computing.  In addition, we establish theoretical results regarding signal tensor recovery, identifiability and  predictive model consistency as well as the algorithm convergence property.

 From a feature extraction perspective, in a sense, the decomposed basis tensor layers in the proposed approach can be viewed as some full-size feature extraction filters analogues to those in the CNN.
 In contrast to the CNN which applies  homogeneous filters over all subjects, the individualized layers in the proposed method  serve as  subject-specific filters  to capture additional  individual-wise characteristics and heterogeneity information, which enables us to effectively capture  heterogeneous outcome-associated  signals  such as the tumor-associated microvesicles (TMVs) in breast cancer imaging data.     \cite{Wang:2017} propose a similar idea for a multiple graph factorization model via decomposing the edge  probability matrix to a population-shared baseline  and an individual deviation. However,  \cite{Wang:2017}'s model is not  applicable to the problem considered in this paper, as their approach focuses on  binary data and assumes a symmetric matrix structure, while we have a multimodal three-way tensor covariate with continuous measurements of pixels.

 Another advantage of the proposed method is that the individualized layer efficiently integrates multimodality  images from the same subject, as it extracts a common structure over different modalities while accounting for modality-specific background noise. Multimodal imaging analysis has drawn great attention in recent years, e.g.,  in cross-modality registration \citep{Maes:1997, Luan:2008} and multiple image fusion \citep{Wang:2006}.  To the best of our knowledge, most existing methods model the inter-modality information either after selecting features from each imaging modality  separately \citep{Zhang:2012,  Liu:2014}, or by combining all modalities into a joint model but assuming independence between modalities  \citep{Hinrichs:2011, Yuan:2012}.
In contrast, through the individualized layers in tensor decomposition, the proposed model integrates multimodality information in both feature extraction and prediction modeling more effectively.

The paper is organized as follows. Section 2 introduces  the main methodology and the background. Section 3 presents the estimation and computation framework.  Section 4 establishes theoretical results.  Section 5 provides simulation studies. Section 6 illustrates an application for human breast cancer optical  imaging data. The last section provides concluding remarks and discussion.

\section{Methodology}
\subsection{Notation and Operations}

Tensor, which also refers  to a multi-dimensional array, plays a central role in the proposed approach.   We start with providing a brief summary of tensor notation.  Extensive
references can be found in the review  reports \citep{Kolda:2006, KoldaandBader:2009}.
A  $D$th-order (or $D$-way)  tensor  is a $D$-dimensional array  $\bm{\mathcal{X}} \in \mathbb{R}^{p_1\times p_2\times \ldots \times p_D}$, where the  dimensions  of a tensor are also known as modes.  We denote  $p_d$ ($1 \le d \le D$) as the marginal dimension of the $d$th mode.
In this paper,  we use the terms way, mode and order interchangeably. The mode-d matricization of  $\bm{\mathcal{X}}$, denoted as $\bm{\mathcal{X}}_{(d)}$,  is defined as a $(p_d \times \prod_{d'\ne d}p_{d'})$-dimensional  matrix, where the $(i_1,\ldots,i_D)$th element of $\bm{\mathcal{X}}$ maps to the $(i_d, j)$th element of $\bm{\mathcal{X}}_{(d)}$ and $j=1+\sum_{d'\ne d}(i_{d'}-1)\prod_{d'' <  d', d''\ne d}p_{d''}$ \citep{KoldaandBader:2009}. Moreover, we denote $\mbox{vec}(\cdot)$ as a vectorizing operation which converts a tensor to a vector, where the element $x_{i_1,\ldots,i_D}$ in a $D$-way tensor $\bm{\mathcal{X}}$ is turned to be the $\left(i_1+\sum_{d=2}^D [(i_d-1)\prod_{j=1}^{d-1}p_j]\right)$th element in the long vector $\mbox{vec}(\bm{\mathcal{X}})$.

Next we  introduce some useful matrix and tensor operations. Given two matrices $\bm{A} \in \mathbb{R}^{m\times n}$ and $\bm{B} \in \mathbb{R}^{p \times q}$, the Kronecker product is an $mp\times nq$ matrix denoted as $\bm{A}\otimes \bm{B}$. If $\bm{A}$ and $\bm{B}$ have the same number of columns $n=q$, then the Khatri-Rao product \citep{Rao:1971} is defined as an $mp\times n$ matrix by $\bm{A} \odot \bm{B}=[\bm{A}_{\cdot 1}\otimes \bm{B}_{\cdot 1} \;  \cdots  \; \bm{A}_{\cdot n}\otimes \bm{B}_{\cdot n}]$, where $\bm{A}_{\cdot k}$ is the $k$th column of $\bm{A}$.  The inner product $\langle \cdot, \cdot \rangle$ of two tensors with the same dimension is defined as $\langle \bm{\mathcal{A}}, \bm{\mathcal{B}} \rangle = \langle \mbox{vec}(\bm{\mathcal{A}}), \mbox{vec}(\bm{\mathcal{B}})\rangle$, which follows that $\langle \bm{\mathcal{A}}, \bm{\mathcal{A}} \rangle= \parallel \bm{\mathcal{A}} \parallel^2_F$, where $\parallel\cdot\parallel_F$ is the Frobenius norm. Moreover, an outer product ``$\circ$'' operating on multiple vectors $\bm{b}^{1} \in \mathbb{R}^{p_1}, \ldots, \bm{b}^{D} \in \mathbb{R}^{p_D}$  creates  a rank-1 $D$-way tensor $\bm{\mathcal{B}}=\bm{b}^{1}\circ \bm{b}^{2}\circ \ldots \circ \bm{b}^{D}$,
where the $(i_1, i_2, \ldots, i_D)$th element of $\bm{\mathcal{B}}$ is defined as $b_{i_1,\ldots,i_D}=b^{1}_{i_1}b^{2}_{i_2}\ldots b^{D}_{i_D}$, and $ \bm{b}^{d}=(b^{d}_1, \ldots, b^{d}_{p_d})'$ is the factor vector at mode $d$.

Consequently, a $D$-way tensor $\bm{\mathcal{B}} \in \mathbb{R}^{p_1\times p_2\times \ldots \times p_D}$ is of rank $R$ if it can be represented as
$ \bm{\mathcal{B}}=\sum_{r=1}^R \bm{b}_r^{1}\circ \bm{b}_r^{2}\circ \ldots \circ \bm{b}_r^{D},$
where $\bm{b}^{d}_r$'s ($r=1,\ldots, R$) are $p_d$-dimensional vectors ($d=1,\ldots,D$).      For ease of notation,  we introduce the Kruskal operator \citep{KoldaandBader:2009}  for a rank-$R$ $D$-way tensor as $\bm{\mathcal{B}}=\bm{B}^{1}\circ\bm{B}^{2}\circ \cdots\circ\bm{B}^{D}=\llbracket\bm{B}^{1}, \bm{B}^{2},\ldots,\bm{B}^{D} \rrbracket$, where $\bm{B}^{d}=[\bm{b}_1^{d}, \bm{b}_2^{d}, \ldots, \bm{b}_R^{d}]\in \mathbb{R}^{p_d \times R}$ denotes the factor matrix at mode $d$.
This  tensor rank decomposition is also known  as  CANDECOMP/PARAFAC  (CP) decomposition,  or Kruskal decomposition \citep{Kruskal:1989}.  An alternative tensor decomposition is  high-order singular value decomposition, also called Tucker decomposition \citep{Tucker:1966}, which decomposes a tensor into a $D$-way core tensor associated with $D$ orthonormal bases-matrices. However, the core tensor in  Tucker decomposition is not guaranteed to be diagonal, and thus the tensor rank is not very clear.  Therefore we adopt the CP decomposition here.

\subsection{Basic Framework and Background}
The sample imaging data is usually a mixture of true signal structure  and random noises.  In general, we assume that
\begin{equation}\label{tensor}
\setlength{\abovedisplayskip}{5pt}
\setlength{\belowdisplayskip}{7pt}
\bm{\mathcal{X}}_i=\bm{\Theta}_i+\bm{\mathcal{N}}_i, \quad i=1,\ldots, N,
\end{equation}
where $\bm{\mathcal{X}}_i \in \mathbb{R}^{p_1\times \cdots \times p_D}$ is the observed  $D$-way sample tensor covariate for the $i$th subject, $\bm{\Theta}_i$  is the true  signal tensor associated with the outcome response, and $\bm{\mathcal{N}}_i$ is the corresponding dimensional noise background. The elements of $\bm{\mathcal{N}}_i$ are independent and identically distributed with  mean of zero and variance of $\sigma^2$.

Let $f(\cdot)$ denote an appropriate feature extraction or transformation mapping on the tensor predictor  $ \bm{\Theta}_i\longmapsto f(\bm{\Theta}_i): \mathbb{R}^{p_1\times \cdots \times p_D} \longrightarrow \mathbb{R}^{q} $.   In a sample version with recovered signal $\hat{\bm{\Theta}}_i$, let $f(\bm{\mathcal{X}}_i)=f(\hat{\bm{\Theta}}_i)$ denote the extracted sample feature.
A general supervised learning stage to link the image tensor covariate $\bm{\mathcal{X}}_i$ and the observed  outcome response $y_i$ can be modeled as
\begin{equation}\label{eq:glm}
\setlength{\abovedisplayskip}{5pt}
\setlength{\belowdisplayskip}{5pt}
\min_{\bm{\beta}} \frac{1}{N}\sum_{i=1}^N \mathcal{L}\left(y_i, f(\bm{\mathcal{X}}_i), \bm{\beta}\right),
\end{equation}
where $\mathcal{L}(\cdot)$ is a selected loss function related to a predictive model  based  on the response types.  For example,  an $L_2$ loss,  $ \mathcal{L}(y_i, f(\bm{\mathcal{X}}_i), \bm{\beta})=\|y_i-f(\bm{\mathcal{X}}_i)^T \bm{\beta}\|^2$,  is usually applied for a linear regression model with a continuous response variable, while for the binary response in  real data,  we can employ a
logistic regression where the corresponding $\mathcal{L}(\cdot)$  refers to the negative loglikelihood function.  Alternatively,  it is also flexible to  utilize  a machine learning model such as  the support vector machine \citep[SVM,][]{Vapnik:1995}   with a hinge loss for a classification problem.  Furthermore,  regularization and penalty  can be also incorporated to the model in (\ref{eq:glm}).

Apparently,  that feature extraction and data transformation is the key part to utilize the imaging information.
One naive transformation method is to directly unfold the imaging tensor $\bm{\mathcal{X}}_i$ to a long vector via  $f(\bm{\mathcal{X}}_i)=\mbox{vec}(\bm{\mathcal{X}}_i)$. However, the number of unknown parameters in (\ref{eq:glm}) using  the  vectorizing covariates is $\prod_{d=1}^{D}p_d$, which is  ultra-high-dimensional and would lead to an unestimable model even applying common regularization methods. In addition, some measurement errors on the imaging data are also incorporated  to the model in (\ref{eq:glm}).  A natural solution to solve these problems  is to employ a dimension reduction technique to recover true signal structure and thus extract important features from the tensor covariates, and then fit the model in (\ref{eq:glm}) based on the  extracted information.  Intuitively, we consider a low-rank approximation for tensor $\bm{\mathcal{X}}_i$ as
\begin{equation*}
%\label{eq:basis}
\setlength{\abovedisplayskip}{2pt}
\setlength{\belowdisplayskip}{5pt}
\bm{\mathcal{X}}_i \approx \sum_{r=1}^R w_{ir}\bm{\mathcal{B}}_r, \quad i=1,\ldots,N,
\end{equation*}
where $\bm{\mathcal{B}}_r$'s ($r=1,\ldots,R$) are normalized $D$-way tensor bases shared  by populations. This rank-$R$ structure can be estimated  by minimizing the difference between the observed imaging and approximated values under the Frobenius norm: $\sum_{i=1}^N \parallel \bm{\mathcal{X}}_i - \sum_{r=1}^R w_{ir}\bm{\mathcal{B}}_r \parallel^2_F$, and thus let $f(\bm{\mathcal{X}}_i)=(\hat{w}_{i1},\ldots,\hat{w}_{iR})'$ be the extracted features, where the $\hat{w}_{ir}$'s are the projection loadings of the $\bm{\mathcal{X}}_i$ onto the estimated bases. However, without specifying any structures on tensor bases, the general  approximation above is still likely to be unestimable and inefficient,  as it is essentially a singular value decomposition problem corresponding to  an $N \times \prod_{d=1}^Dp_d$-dimensional matrix for the vectorized data.

\subsection{Individualized Multilayer Tensor Decomposition}

In this subsection, we propose a novel individualized multilayer tensor decomposition framework for true signal recovery and  feature extraction. Following the model in (\ref{tensor}),  we have a $D$-way tensor covariate $\bm{\mathcal{X}}_i \in \mathbb{R}^{p_1\times \cdots \times p_D} $ for each individual ($i=1,\ldots,N$), and  let $\bm{\mathcal{X}}=[\bm{\mathcal{X}}_1 \; \cdots \; \bm{\mathcal{X}}_N] \in \mathbb{R}^{N\times p_1\times  \cdots \times p_D}$  denote the  aligned higher-order tensor covariate,  $\bm{\Theta}=[\bm{\Theta}_1 \; \cdots \; \bm{\Theta}_N]$ denote the aligned true signal tensor, and $\bm{\mathcal{N}}=[\bm{\mathcal{N}}_1 \; \cdots \; \bm{\mathcal{N}}_N]$ denote the corresponding dimensional noise tensor.   In general,  the observed tensor covariate is mixed with a true signal structure  and some random noises:
$
\bm{\mathcal{X}}=\bm{\Theta}+\bm{\mathcal{N}}.
$

To motivate our model,  we first start with considering a low-rank structure modeling of the grand signal tensor $\bm{\Theta}$ following a CP decomposition:
\begin{equation}\label{eq:hocpd}
\setlength{\abovedisplayskip}{5pt}
\setlength{\belowdisplayskip}{7pt}
\bm{\Theta}=\bm{W}\circ\bm{B}^{1}\circ\cdots\circ\bm{B}^{D}
=\sum_{r=1}^R \bm{W}_{\cdot r}\circ \bm{B}^{1}_{\cdot r}\circ\cdots\circ\bm{B}^{D}_{\cdot r},
\end{equation}
where $\bm{B}^{d}_{\cdot r}$ ($r=1,\ldots, R$) denotes the $r$th column vector of mode-$d$ factor matrix $\bm{B}^{d} \in \mathbb{R}^{p_d \times R}$ ($d=1,\ldots,D$), and $\bm{W}=[\bm{w}_1,\ldots,\bm{w}_N]' \in \mathbb{R} ^{N \times R}$ is the factor matrix corresponding to the dimension of individuals, and $\bm{w}_i=(w_{i1},\ldots,w_{iR})'$ is the loading vector associated to the $i$th individual.
To avoid the indeterminacy from scaling, we let $ \bar{\bm{B}}_{\cdot r}^{d} = \frac{1}{\|\bm{B}_{\cdot r}^{d}\|}\bm{B}_{\cdot r}^{d}$ be the normalized factor vector at mode $d$. We further Let $\bm{\mathcal{B}}_r= \bar{\bm{B}}_{\cdot r}^{1}\circ \bar{\bm{B}}_{\cdot r}^{2}\circ \ldots \circ \bar{\bm{B}}_{\cdot r}^{D}$ and $\bar{w}_{ir}=w_{ir}\prod_{d=1}^D \|\bm{B}_{\cdot r}^{d}\|$, and thus $\|\bm{\mathcal{B}}_r\|_F=1$ (see supplementary material A.1 for  details).
Then each individual tensor can be represented as
\begin{equation}
\label{eq:hocpdb}
\setlength{\abovedisplayskip}{7pt}
\setlength{\belowdisplayskip}{7pt}
\bm{\mathcal{X}}_i=\bm{\Theta}_i+\bm{\mathcal{N}}_i=\sum_{r=1}^R \bar{w}_{ir}\bm{\mathcal{B}}_r+ \bm{\mathcal{N}}_i,
\end{equation}
where $\bm{\mathcal{B}}_r$'s serve as population-wise rank-$1$ basis tensors, and thus the individual-specific feature based on  the signal tensor $\bm{\Theta}_i$  is pooled as $f(\bm{\Theta}_i)=\bar{\bm{w}}_i=(\bar{w}_{i1},\ldots,\bar{w}_{iR})'$.  In the sample version, the extracted feature based on the observed sample  covariates is $f(\bm{\mathcal{X}}_i) = f(\hat{\bm{\Theta}}_i)=\hat{\bar{\bm{w}}}_i$,  where the estimated  signals $\hat{\bm{\Theta}}_i$'s  as well as the $\hat{\bm{w}}_i$'s are obtained   by minimizing an $L_2$-loss: $ \|\bm{\mathcal{X}} -\bm{\Theta} \|^2_F$ under the model in (\ref{eq:hocpd}).   We refer this model to  the higher-order CP decomposition (HOCPD) model.    The  HOCPD method   is powerful  for reducing  the tensor covariates' dimensionality,  as the  feature dimension  is effectively reduced from $\prod_{d=1}^D p_d$  to $R$ for the supervised learning in (\ref{eq:glm}); however,  it  highly depends on the low-rank assumption on the grand signal tensor,  which could fail to capture complex tensor data  information  if there is  significant heterogeneous variation  arising from unique individuals.  Moreover, it is  infeasible to capture  heterogeneity simply by increasing the rank in an HOCPD model, as it might lead to non-identifiability and unstable model estimation.

In the following, we propose an individualized multilayer tensor learning (IMTL) method to incorporate heterogeneous structures to tensor decomposition.  For the $i$th individual, we assume
\begin{equation}\label{IMM}
\setlength{\abovedisplayskip}{5pt}
\setlength{\belowdisplayskip}{7pt}
\bm{\Theta}_i=\sum\limits_{r=1}^R \bar{w}_{ir} \bm{\mathcal{B}}_r  + \bm{\mathcal{S}}_i, \quad
\mbox{ s.t.} \quad \big\langle \bm{\mathcal{B}}_r, \bm{\mathcal{S}}_i \big\rangle=0,\; 1 \le r \le R,
\end{equation}
where $\bm{\mathcal{B}}_r$'s are the  population-shared rank-1 bases  analogous to those in (\ref{eq:hocpdb}),  and $\bm{\mathcal{S}}_i=\bm{s}_i^{1}\circ \bm{s}_i^{2}\circ \ldots \circ \bm{s}_i^{D}$ is an individualized rank-1 structure. With normalized vectors $\bar{\bm{s}}_i^{d}$'s and $u_i=\prod_{d=1}^D \|\bm{s}^d_i\|$, we can rewrite  $\bm{\mathcal{S}}_i=u_i\bar{\bm{\mathcal{S}}}_i$ where $\bar{\bm{\mathcal{S}}}_i=\bar{\bm{s}}_i^{1}\circ \bar{\bm{s}}_i^{2}\circ \ldots \circ \bar{\bm{s}}_i^{D}$.   Given observed  sample tensor covariates,  the latent parameters in model (\ref{IMM}) are  estimated through
$$
\setlength{\abovedisplayskip}{3pt}
\setlength{\belowdisplayskip}{5pt}
 \min_{\bm{\bm{w}}_i, \bm{\mathcal{B}}_r, \mu_i, \bar{\bm{\mathcal{S}}}_i} \sum_{i=1}^N \|\bm{\mathcal{X}}_i -\sum\limits_{r=1}^R \bar{w}_{ir} \bm{\mathcal{B}}_r  - \mu_i \bar{\bm{\mathcal{S}}}_i \|^2_F, \quad
\mbox{ s.t.} \quad \big\langle \bm{\mathcal{B}}_r, \bar{\bm{\mathcal{S}}}_i \big\rangle=0,\; 1 \le r \le R.
$$
The orthogonal constraint is imposed  to ensure the identifiability between the  homogeneous layers and the heterogeneous  layers, and also to improve the computational stability.   Based on the recovered signal $\hat{\bm{\Theta}}_i$,  the extracted individual feature is $f(\hat{\bm{\Theta}}_i)=(\hat{\bar{\bm{w}}}'_i, \hat{u}_i, \{\hat{\bar{\bm{s}}}^{'d}_i\}^D_{d=1})'$, containing the weights of the population-shared  layers $\hat{\bar{\bm{w}}}_i=(\bar{w}_{i1}, \ldots,\bar{w}_{iR})'$, the weight of the individualized layer $\hat{\mu}_i$, and the decomposed factors of the individualized layers  $\hat{\bm{\bar{s}}}^{d}_i$'s, respectively.

In contrast to the HOCPD  in (\ref{eq:hocpdb}), each signal tensor  $\bm{\Theta}_i$ in (\ref{IMM}) is characterized  by two different layers, one consisting of a linear combination of  homogeneous  structures ($\bm{\mathcal{B}}_r$'s), and another containing  an individualized structure ($\bm{\mathcal{S}}_i$)  capturing the   heterogeneity of individual features.
 Intuitively, the individualized layer  $\bm{\mathcal{S}}_i$ is the ``best'' rank-1 structure extracted from the ``residual'' tensor of a higher-order decomposition.  This additional  layer enables us to capture the individual-specific  information of the tensor covariate which is not  characterized  by a   population structure $\sum\limits_{r=1}^R \bar{w}_{ir} \bm{\mathcal{B}}_r$,    while the rank-1 structure in (\ref{IMM}) avoids overfitting on the individualized layers.

Additionally, In practice,  the informative heterogeneous  signals could be relatively weak compared to  the population-wise background and noise.   For example, in real data, the important biomarker TMVs are discrete and  random spatially distributed, while the signal strength is also quite weak.  Therefore, they are very likely to be washed out  if only low-rank homogeneous bases are used.  In some sense, the homogeneous bases in the population layer serve  as some full-size  filters to remove those homogeneous variation, so the subject-specific signals which deviate from the low-rank homogeneous background can be captured.   Consequently, the population layer and the individual layer  effectively coordinate to extract informative features from sample tensor covariates and thus improve model prediction power.

\subsection{Multilayer Tensor Reconstruction for Multimodality Data}

Multimodality imaging data is widely adopted and has drawn great attentions in recent years.  In our motivating application,  optical imaging  produces multiple images using different wavelengths of light in one examination.
Although  a single modality model is applicable,  there is a critical need  to integrate  all information collected from multiple modalities, so important features associated with clinical outcomes can be extracted more effectively and the prediction power can be greatly enhanced.
In the following,  we develop a multimodality  imaging  tensor model which  extends  the  individualized multilayer approach in (\ref{IMM}) to incorporate different sources of modality information.

We consider an M-modality  tensor covariate $\bm{\Theta}=[\bm{\Theta}^{(1)} \; \cdots \; \bm{\Theta}^{(M)}]$, where  $\bm{\Theta}^{(m)} \in \mathbb{R}^{N \times p_1\times \ldots \times p_D}$  ($1 \le m \le M$). For a single modality $m$,  we propose a multilayer tensor decomposition
\[
\setlength{\abovedisplayskip}{3pt}
\setlength{\belowdisplayskip}{5pt}
\bm{\Theta}^{(m)}=\bm{W}\circ\bm{B}^{(m),1}\circ\cdots\circ\bm{B}^{(m),D}
+\bm{\mathcal{S}},
\]
where $\bm{B}^{(m),d}$'s are modality-specific factors, and   $\bm{\mathcal{S}}=[\bm{\mathcal{S}}_1 \; \bm{\mathcal{S}}_2
\; \ldots \; \bm{\mathcal{S}}_N] \in \mathbb{R}^{N \times p_1\times \ldots \times p_D}$ contains heterogenous layers.   With normalized factors,  each individual tensor can be rewritten as
\begin{equation}
\label{eq:MIMM}
\setlength{\abovedisplayskip}{7pt}
\setlength{\belowdisplayskip}{8pt}
\bm{\Theta}^{(m)}_i = \sum\limits_{r=1}^{R} \bar{w}^{(m)}_{ir} \bm{\mathcal{B}}^{(m)}_r +  \mu_i\bar{\bm{\mathcal{S}}}_i,    \quad \mbox{s.t.} \quad \big\langle \bm{\mathcal{B}}^{(m)}_r, \bar{\bm{\mathcal{S}}}_i \big\rangle=0,\;  1 \le r \le R,
\end{equation}
where $\{\bm{\mathcal{B}}_r^{(m)} = \bar{\bm{B}}^{(m),1}_{\cdot r}\circ\cdots\circ\bar{\bm{B}}^{(m),D}_{\cdot r}\}_{r=1}^R$ is the set of bases for the $m$th modality, and $\bm{\mathcal{S}}_i=\mu_i\bar{\bm{\mathcal{S}}}_i$ is an individualized layer shared by different modalities $\bm{\Theta}^{(m)}_i$ ($m=1,\ldots,M$) of the $i$th subject.
Similarly, the proposed model in (\ref{eq:MIMM}) can be  estimated by minimizing a sum of squared loss:  $ \sum_{m=1}^{ M} \parallel \bm{\mathcal{X}}^{(m)} -\bm{\Theta}^{(m)} \parallel^2_F$  with the constraints.  Let $\bm{\Theta}_i = [\bm{\Theta}^{(1)}_i \ldots \bm{\Theta}^{(M)}_i]$ be the multimodality tensor signal for the $i$th subject,  and denote $\hat{\bm{\Theta}}_i$ as the sample  estimator based on (\ref{eq:MIMM}).  The extracted subject-wise feature is  $f(\hat{\bm{\Theta}}_i)= (\hat{\bar{\bm{w}}}_i, \hat{\mu}_i, \{ \hat{\bar{\bm{s}}}^{d}_i\}_{d=1}^D)'$, containing the weights of the modality specific layers $\hat{\bar{\bm{w}}}_i = \{\hat{\bar{w}}^{(m)}_{ir}\}^{1 \le m \le M}_{ 1\le r \le R}$, the weight of the individualized layer $\hat{\mu}_i$, and the decomposed factors of the individualized layer $\hat{\bm{\bar{s}}}^{d}_i$'s, respectively.

For the proposed multimodality model in (\ref{eq:MIMM}), the individualized layer $\bm{\mathcal{S}}_i$ is a common low-rank structure capturing the individual-specific spatial pattern shared by different modalities. Similarly, for a given subject,  if we consider removing the modality-specific layers  $\sum\limits_{r=1}^{R} \bar{w}^{(m)}_{ir} \bm{\mathcal{B}}^{(m)}_r$ from each modality $\bm{\Theta}_i^{(m)}$ first, and then aligning     the ``residuals''  from multiple modalities to a higher-order tensor ($\mathbf{R}^{p_1 \times \cdots \times p_D \times M}$),  then the individualized layer can be interpreted as a tensor slice along the modality mode of  the ``best''  rank-1 structure extracted from the higher-order residual tensor.

\begin{figure}[H]
    \centering
    \includegraphics[width=1\textwidth, height=10.5cm]{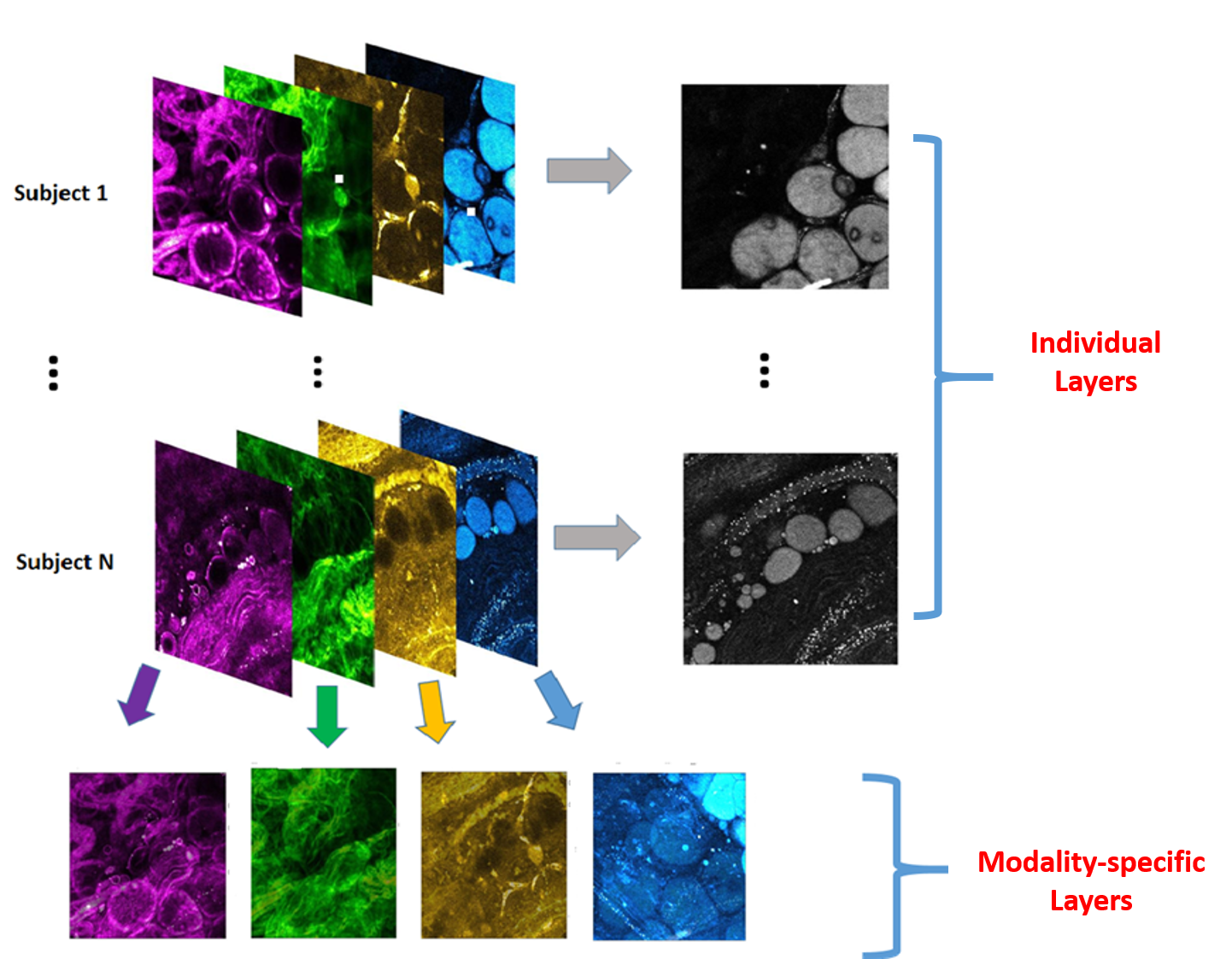}
\caption {An illustration of the individualized layers and the modality-specific layers for  four-modality optical images of the breast cancer tissues.}
\label{im:ml}
\end{figure}

In addition to characterizing the heterogeneous structures analogues to the singlemodality case, the individualized layer  also makes it feasible to integrate the information of  multiple modalities effectively.  Although the individual-specific signals are randomly distributed across different individuals,  their spatial pattern is shared by different  modalities from the same individual. Therefore, the individualized layer extracted over multiple modalities  aggregates the information regarding  their common structure  and thus captures this  individual-specific feature more accurately.   Figure \ref{im:ml} provides an illustration of the individualized layers and the modality-specific layers on the four-modality  breast cancer images.

Alternatively,  we can  simply align the four modalities  together to a higher-order tensor predictor with a size of $p_1 \times ... \times  p_D \times M$, and then apply the proposed  singlemodality model in (\ref{IMM}).
However, this is inadequate to capture the variation among different modalities.  In the following,  we compare  the  singlemodality model  in (\ref{IMM}) with the integrated tensor and   multimodality model in (\ref{eq:MIMM}).  Let  $\bm{\Theta}_i=[\bm{\Theta}^{(1)}_i, \ldots, \bm{\Theta}^{(M)}_i ] \in \mathbf{R}^{p_1\times \cdots \times p_D \times M}$ denote  the multimodality integrated tensor.
With a singlemodality model,  we have
$$
\setlength{\abovedisplayskip}{3pt}
\setlength{\belowdisplayskip}{5pt}
 \bm{\Theta}_i=\sum_{r=1}^R w_{ir}\bm{b}^1_r\circ\cdots\circ\bm{b}^D_r\circ\bm{b}_r^{modal} + \bm{s}^1_i\circ\cdots\circ\bm{s}^D_i\circ\bm{s}_i^{modal} ,
 $$
 where $\bm{b}_r^{modal}=(b^{modal}_{1r}, \ldots,b^{modal}_{Mr})'$ and $\bm{s}_i^{modal}=(s^{modal}_{i1}, \ldots,b^{modal}_{iM})'$ are both basis vectors on the mode of modality.
Hence, for the $m$th modality ($m=1,2,3,4$), the corresponding tensor slice ($p_1\times \cdots \times p_D$-dimensional tensor) can be written as
$$
\setlength{\abovedisplayskip}{3pt}
\setlength{\belowdisplayskip}{5pt}
\bm{\Theta}^{(m)}_i=\sum_{r=1}^R w_{ir}b^{modal}_{mr}\bm{b}^1_r\circ\cdots\circ\bm{b}^D_r+s^{modal}_{im}\bm{s}^1_i\circ\cdots\circ\bm{s}^D_i.
$$
This implies that the population-shared  basis layers $\bm{\mathcal{B}}_r=\bm{b}^1_r\circ\cdots\circ\bm{b}^D_r$ ($1\le r \le R$) are the same across all modalities up to scaling and permutation.  In contrast,  the multimodality model in (\ref{eq:MIMM}) allows different modality-specific layers $\bm{\mathcal{B}}^{(m)}_r$ across different modalities  to capture the modality-wise heterogeneity. In fact, the singlemodality  model on the integrated data is a special case of the multimodality model (\ref{eq:MIMM}),  with additional constraints on all modalities sharing the same set of population bases.

\section{Computation}
In this section, we address the estimation problem of multilayer decomposition in (\ref{eq:MIMM}) for the proposed IMTL model. In contrast to traditional CP decomposition, incorporating modality layers and  individual layers of the proposed method significantly increases the computation cost, and conventional algorithms for tensor decomposition are not necessarily  scalable in our situation. Therefore, we propose a  bi-level block improvement algorithm which alternately updates different layers and apply a maximum block improvement (MBI) strategy for the estimation of each layer.

The proposed model in (\ref{eq:MIMM}) yields a constrained optimization  requiring  orthogonality between the modality layers and the individual layers. Hence,  we employ a penalization method \citep{Nocedal:2006} to achieve orthogonality constrained optimization, that is, to minimize the objective function
\begin{equation}
\label{obj}
 \setlength{\abovedisplayskip}{5pt}
\setlength{\belowdisplayskip}{7pt}
  L\left(\bm{W},\{\bm{B}^{(m),d}\}_{m's, d's},\; \{\bm{s}^d_i\}_{i's ,d's} \;|\bm{\mathcal{X}}\right)=\sum_{m=1}^M \|  \bm{\mathcal{X}}^{(m)}  - \bm{\Theta}^{(m)} \|^2_F + \lambda_{s} \sum_{m,i,r} \langle \bm{\mathcal{B}}^{(m)}_r, \bm{\mathcal{S}}_i \rangle^2.
\end{equation}
Note that the orthogonal property of the decomposed layers is not critical here as it only serves to ensure  identifiability. Although the penalization method cannot guarantee an exact orthogonality of the estimated layers, it is sufficient for pursuing the layers' identifiability in practice.

The objective function in (\ref{obj}) is convex and differentiable with respect to each separate block of parameters, which makes it applicable to apply a block-wise updating procedure. However, the traditional tensor decomposition algorithms, e.g., alternating least squares \citep[ALS,][]{Carroll:1970} and block relaxation algorithm \citep{Leeuw:1994, Lange:2010}, are nearly infeasible and unscalable due to the large number ($(M+N)D+1$) of blocks, and updating one block at each iteration could lead to  poor convergence performance.  In addition, the estimation of different blocks (layers) utilizes different parts of the data, for example, $\bm{B}^{(m),d}$'s are estimated within each modality and $\bm{s}_i^{d}$'s are estimated within each subject. This allows for alternately estimating different layers, and makes parallel computing feasible,  which significantly improves the computational scalability.  Finally, in general, for $D>2$,  the ALS-type cyclical algorithm is not guaranteed to converge  to a stationary point \citep{Chen:2010}.   Therefore, we propose a bi-level algorithm, that is,   within the estimation of each layer, we apply the MBI strategy \citep{Chen:2010} in estimating specific blocks, and then we update each layer alternately.

Specifically, given the individual-layer factors, we let $\bm{\mathcal{X}}^{(m)} = \bm{\mathcal{X}}^{(m)} - \hat{\bm{\mathcal{S}}}$ and alternately update blocks of $\bm{W}$ and $\{\bm{B}^{(m),d}\}$'s; that is,
\begin{equation}
\label{als:W}
\setlength{\abovedisplayskip}{5pt}
\setlength{\belowdisplayskip}{7pt}
\hat{\bm{W}}^{[new]}=\argmin_{\bm{W}} \sum_{m=1}^{M} \left\| \tilde{\bm{\mathcal{X}}}^{(m)}_{(1)}-  \bm{W} \big( \hat{\bm{B}}^{(m),D}\odot \cdots \odot \hat{\bm{B}}^{(m),1} \big)^T  \right\|^2_F,
\end{equation}
where $\tilde{\bm{\mathcal{X}}}^{(m)}_{(1)}$ is the  mode-1 matricization.  The sub-optimization  turns to be a ridge-type problem and thus has a unique explicit solution.  Next, given $\hat{\bm{W}}$, we update $\{\bm{B}^{(m),d}\}_{d's}$ through parallel computing across different modalities ($m=1,\ldots, M$), and apply the MBI strategy. Let
\begin{equation*}
\setlength{\abovedisplayskip}{5pt}
\setlength{\belowdisplayskip}{7pt}
%\resizebox{0.93\hsize}{!}
%{$
 L^{(m)}\big(\bm{B}^{(m),d}| \{\bm{B}^{(m),d'}\}_{d' \ne d}\big) =\left\| \tilde{\bm{\mathcal{X}}}^{(m)}_{(d+1)}- \bm{B}^{(m),d} \big(  \bm{B}_{-d}^{(m)} \odot \hat{\bm{W}}  \big)^T  \right\|^2_F+\lambda_s \sum_{i} \Big\langle  \bm{B}^{(m),d}, \; \hat{\bm{\mathcal{S}}}_{i,(d)} \bm{B}_{-d}^{(m)} \Big\rangle^2
%$}
\end{equation*}
be the reduced objective function with respect to $\bm{B}^{(m),d}$ while fixing all other blocks,  where $\bm{B}_{-d}^{(m)} = \bm{B}^{(m),D}\odot \cdots \bm{B}^{(m),d+1} \odot \bm{B}^{(m),d-1} \cdots \odot \bm{B}^{(m),1}$ and $\hat{\bm{\mathcal{S}}}_{i,(d)}$ is the mode-d matricization of $\hat{\bm{\mathcal{S}}}_{i}$. Given the latest iterate $\{\bm{B}^{(m),1, [t]},\ldots,\bm{B}^{(m),D,[t]}\}$, we calculate
\begin{equation}
\label{als:B}
\setlength{\abovedisplayskip}{5pt}
\setlength{\belowdisplayskip}{7pt}
\begin{split}
&\bm{B}^{*(m),d}= \argmin_{\bm{B}^{(m),d}} L^{(m)}(\bm{B}^{(m),d}| \{\bm{B}^{(m),d',[t]}\}_{d' \ne d}),\\
&d_m^*=\arg\min_{1 \le d \le D} L^{(m)}(\bm{B}^{*(m),d}| \{\bm{B}^{(m),d',[t]}\}_{d' \ne d}),
\end{split}
\end{equation}
and update only the block $\bm{B}^{(m),d,[t]}$ at the mode $d=d_m^*$   for  maximum improvement.

Similarly, given the modality-layer factors, the individualized layers can be estimated within each subject using  parallel computing.  For each subject $i$, let $\bar{\bm{\mathcal{X}}}^{(m)}_i=\bm{\mathcal{X}}^{(m)}_i - \sum\limits_{r=1}^R \hat{w}_{ir}\hat{\bm{\mathcal{B}}}^{(m)}_r$, and let
\begin{equation*}
\setlength{\abovedisplayskip}{5pt}
\setlength{\belowdisplayskip}{7pt}
 L_i\big(\bm{s}_i^{d}\;|\; \{\bm{s}_i^{d'}\}_{d' \ne d}\big) =\sum_{m=1}^M \|  \bar{\bm{\mathcal{X}}}^{(m)}_{i, (d)}  -  \bm{s}_i^d (\bm{S}_{i,-d})^T \|^2_F+\lambda_s \sum_{m,r} \langle  \bm{s}_i^d, \; \hat{\bm{\mathcal{B}}}^{(m)}_{r,(d)} \bm{S}_{i,-d} \rangle^2
\end{equation*}
be the reduced objective function with respect to $\bm{s}_i^{d}$,  where $\bm{S}_{i,-d} = \bm{s}_i^1 \odot \cdots  \bm{s}_i^{d-1} \odot \bm{s}_i^{d+1} \cdots \odot \bm{s}_{i}^{D} $ and $\hat{\bm{\mathcal{B}}}^{(m)}_{r,(d)}$ is the mode-d matricization. Given the latest iterate $\{\bm{s}_i^{1, [t]},\ldots,\bm{s}_i^{D,[t]}\}$, calculate
\begin{equation}
\label{als:S}
\setlength{\abovedisplayskip}{5pt}
\setlength{\belowdisplayskip}{7pt}
\bm{s}_i^{*d}= \argmin_{\bm{s}_i^{d}} L_i(\bm{s}_i^{d}| \{\bm{s}_i^{d',[t]}\}_{d' \ne d}),\quad \text{and} \quad
d_i^{*}=\arg\min_{1 \le d \le D} L_i(\bm{s}_i^{*d}| \{\bm{s}_i^{d',[t]}\}_{d' \ne d}),
\end{equation}
and update only mode-$d_i^*$ block.  The detailed algorithm is summarized in following Algorithm \ref{algr}.
\begin{algorithm}[h]
\caption{A Bi-level Block Improvement Algorithm with Parallel Computing}
\label{algr}
1.(\emph{Initialization}) Set $t=1$ and the tuning parameter $\lambda_s$. Set initial values for $\bm{W}^{[0]}$, $\bm{B}^{(m),d, [0]}$'s ($1 \le m \le M; 1 \le d \le D$). Set $\bm{s}_i^{d,[0]}=\bm{0}$ ($1 \le i \le N; 1 \le d \le D$). \\
\\
2.(\emph{Modality layers}) At the $t$th iteration, given $\{\bm{s}_i^{d,[t-1]}\}'s$, estimate $\bm{W}^{[t]}$ and  $\{\bm{B}^{(m),d, [t]}\}'s$. \\
\vspace{-5mm}
   \begin{enumerate}
     \item[(i)] Set $\bm{W}^{[t]} \leftarrow \bm{W}^{[t-1]}$, $\bm{B}^{(m),d, [t]} \leftarrow \bm{B}^{(m),d, [t-1]}$ ($m$'s, $d$'s).
     \item[(ii)] Fixing all $\bm{B}^{(m),d, [t]}$'s,  update $\bm{W}^{[t]_{new}}$ through (\ref{als:W}).
     \item[(iii)] Fixing $\bm{W}^{[t]}$, for each modality $m$, a) calculate $d_{m}^*$'s and $\bm{B}^{*(m),d}$'s based on  (\ref{als:B});  b) assign $\bm{B}^{(m),d,[t]_{new}} \leftarrow \bm{B}^{*(m),d}$ if $d=d_{m}^*$, and $\bm{B}^{(m),d,[t]_{new}} \leftarrow \bm{B}^{(m),d,[t]}$ if  $d \ne d_{m}^*$ .
     \item[(iv)] Stop iteration if $\frac{1}{NR}\|\bm{W}^{[t]_{new}}{-}\bm{W}^{[t]}\|_F^2 {+} \frac{1}{MR\prod_{d=1}^D p_d} \sum_{m,D} \|\bm{B}^{(m),d,[t]_{new}}{-}\bm{B}^{(m),d,[t]}\|_F^2  {\le} 10^{-4}$, otherwise assign $\bm{W}^{[t]}{\leftarrow}\bm{W}^{[t]_{new}}$,  $\bm{B}^{(m),d,[t]}{\leftarrow}\bm{B}^{(m),d,[t]_{new}}$ ($m$'s, $d$'s), and go to Step 2(ii).
   \end{enumerate}
3. (\emph{Individualized layers}). Given $\bm{W}^{[t]}$, $\bm{B}^{(m),d, [t]}$'s, for individual $i$ ($1\le i \le N$), estimate $\bm{s}_i^{d,[t]}$'s.\\
\vspace{-5mm}
   \begin{enumerate}
     \item[(i)] Set $\bm{s}_i^{d,[t]} \leftarrow \bm{s}_i^{d,[t-1]}$ ($d$'s).
     \item[(ii)] a) Calculate $d_i^*$'s and $\bm{s}_i^{*d}$'s based on  (\ref{als:S}); b) assign $\bm{s}_i^{d,[t]_{new}} {\leftarrow} \bm{s}_i^{*d}$ if $d=d_i^*$, and $\bm{s}_i^{d,[t]_{new}} {\leftarrow} \bm{s}_i^{d,[t]}$ if $d\ne d_i^*$.
     \item[(iii)] Stop iteration if $\frac{1}{\prod_{d=1}^D p_d} \sum_{d} \|\bm{s}_i^{d,[t]_{new}} {-} \bm{s}_i^{d,[t]}\|_F^2  {\le} 10^{-4}$, otherwise assign $\bm{s}_i^{d,[t]}{\leftarrow}\bm{s}_i^{d,[t]_{new}}$, and go to Step 3(ii).
   \end{enumerate}
4. (\emph{Stopping Criterion}) Stop if $\frac{\|\bm{\Theta}^{[t]}{-}\bm{\Theta}^{[t-1]}\|_F^2 }{NM\prod_{d=1}^{D}p_d} {\le} 10^{-3}$, otherwise set $t {\leftarrow} t+1$ and go to Step 2.
\end{algorithm}

For any objective function $f(\bm{A}_1,\ldots,\bm{A}_n)$ with blocks of  parameters $\{\bm{A}_i\}_{i=1}^n$, a point of $(\bm{A}^*_1,\ldots,\bm{A}^*_n)$ in the parameter space is defined as a block-wise stationary point of $f(\cdot)$  if for any block, there is
$
\bm{A}^*_i = \arg\min_{\bm{A}_i} f(\bm{A}^*_1,\ldots, \bm{A}^*_{i-1}, \bm{A}_{i}, \bm{A}^*_{i+1},\ldots,\bm{A}^*_n).
$

Let $\{\bm{W}, \bm{B}^{(m),d}, \bm{s}^d_{i}\}_{m,i,d}$ be an element-wise collection of all parameters and  let $\Omega$ be the corresponding parameter space. We provide a global convergence property of Algorithm \ref{algr} as follows.
\begin{lemma}
\label{lm:conv}
Assume $\Omega$ is compact, then any accumulative point of the iterations from Algorithm \ref{algr}, say $\{\bm{W}^*, \bm{B}^{*(m),d}, \bm{s}^{*d}_{i}\}_{m's,i's,d's}$, is a block-wise stationary point of the objective function in (\ref{obj}).
\end{lemma}

In general,  multiple initializations are suggested to obtain a sound optimum in (\ref{obj}). The normalization on modes' factor vectors can be performed after the last iteration.  Given the estimated tensor signals $\hat{\bm{\Theta}}_i$'s,  for binary response,  we employ an $L_1$-penalized logistic regression model  with the extracted sample features $f(\hat{\bm{\Theta}}_i)$'s in (\ref{eq:glm}) for prediction.  The rank of the population-shared layers and the other  tuning parameter are selected based on a grid search to minimize the prediction errors on the validation set or via a cross-validation method.   More details and discussion about  tuning procedure and the prediction on a new subject are provided in the supplementary material.

\section{Theory}

In this section, we develop  the theoretical framework for the proposed model regarding both tensor signal recovery and prediction modeling. First, we introduce conditions to ensure  identifiability of the proposed multilayer tensor modeling. Furthermore, we provide an accurate  error bound for the recovered signal tensor and show that the estimated signal tensor converges to the true one.  Finally, we present the theoretical results of the supervised learning stage to evaluate  prediction performance.  All proofs are provided in supplementary materials Section A.

We first address the tensor modeling identifiability issue before establishing the statistical property, which is critical for tensor representation.  In particular, we provide sufficient conditions to achieve identifiable layers in the proposed multilayer tensor model, which can be verified easily in practice. For ease of notation, the following discussion focuses on the single modality model while the presented conditions can be easily extended to the multimodality model.

In the proposed framework,  potential unidentifiability could occur in the multi-layer CP decomposition of tensor predictors in (\ref{IMM}), which can be attributed to three aspects. The first two indeterminacies arise from scaling and permutation, and  the last aspect is the non-uniqueness of the CP decomposition for a tensor, that is, the possibility that more than one combination of population layers and individualized layers can lead to the underlying true image tensor.
In the proposed model, the scaling indeterminacy, which refers to  possible rescaling over different modes' factor vectors for each layer, is eliminated by imposing a unit-norm constraint on parameterization, that is, $\|\bar{\bm{B}}^{d}_{\cdot r} \|_2 =1$ and $\|\bar{\bm{s}}^{d}_i \|_2 =1$.  Moreover, the permutation indeterminacy refers to the arbitrary reordering of the population bases. To address this point, we could align the population bases according to a descending order of the first element of mode-1 factor vectors, that is,
$
\bar{\bm{B}}^{1}_{11} \ge \bar{\bm{B}}^{1}_{12}  \ge \cdots \ge \bar{\bm{B}}^{1}_{1R}.
$

After controlling the scaling and the permutation, in general, the CP decomposition of a tensor  could still  be non-unique,  due to the possibility of multiple combinations of rank-one tensors in decomposition.   Although various identifiability conditions have been presented for a conventional CP decomposition \citep{Kruskal:1989,  Sidiropoulos:2000, KoldaandBader:2009},  they are infeasible here as  all of them require checking each individual tensor $\bm{\Theta}_i$ ($1\le i \le N$) separately, which is not effective in practice, especially when the sample size $N$ is increasing.

In the following, we provide a much weaker sufficient condition based on the integrated higher-order tensor without imposing any additional constraints on parameterization.  We first introduce the concept of a $k$-rank of a matrix according to \citep{Kruskal:1989}. Specifically,  the $k$-rank of a matrix $\bm{A}$, denoted as $\mathcal{K}_{\bm{A}}$  is defined as
$
\mathcal{K}_{\bm{A}}= \max \{ k:\;\text{any $k$ columns of $\bm{A}$  are linearly independent} \}.
$
Let $\bm{\Theta}_{[1:n]}$ denote an integrated $(D+1)$-way tensor combining $n$ individual tensors.  Without loss of generality, we assume $\bm{\Theta}_{[1:n]}=[\bm{\Theta}_1 \; \cdots \; \bm{\Theta}_n]$. There is a $(R+n)$-rank representation for the integrated tensor:
\[
\setlength{\abovedisplayskip}{2pt}
\setlength{\belowdisplayskip}{7pt}
\bm{\Theta}_{[1:n]} =  \sum_{r=1}^R \bm{w}^{[1:n]}_{r} \circ \bm{B}^{1}_{\cdot r} \circ \cdots \circ \bm{B}^{D}_{\cdot r} + \sum_{i=1}^n \tilde{\bm{u}}^{[1:n]}_{i} \circ \bm{s}_i^{1} \circ \cdots \circ \bm{s}^{D}_i,
\]
where $\bm{w}^{[1:n]}_{r}=(w_{1r},\ldots,w_{nr})'$ and $\tilde{\bm{u}}^{[1:n]}_{i}=(\underbrace{0, \ldots, 0}_{1,\ldots,i-1},\underbrace{1}_{i},\underbrace{0, \ldots, 0}_{i+1,\ldots, n})'$. For $1 \le d \le D$, let
$\tilde{\bm{B}}^{d}_{[1:n]}=[\bm{b}^{d}_1 \; \cdots \; \bm{b}^{d}_R \; \bm{s}^{d}_1 \; \cdots \; \bm{s}^{d}_n]$ denote the mode-d factor matrix  for integrated  tensor $\bm{\Theta}_{[1:n]}$.
We have the following proposition providing a sufficient condition for the identifiability of the multi-layer decomposition in (\ref{eq:MIMM}).

\begin{prop}\label{prop:idt2}
If there exist $n$  individual tensors ($2 \le n \le N $) in $\{ \bm{\Theta}_i, 1\le i \le N\}$, such that,
\[
\setlength{\abovedisplayskip}{2pt}
\setlength{\belowdisplayskip}{5pt}
\sum_{d=1}^{D} \mathcal{K}_{\tilde{\bm{B}}^{d}_{[1:n]}} \ge 2R + n + D
\]
holds for the integrated high-order tenor $\bm{\Theta}_{[1:n]}$, then the multi-layer decomposition in (\ref{IMM}) is unique given the unit-norm and the ordering constraints on the factor vectors.
\end{prop}

In the proof of Proposition \ref{prop:idt2}, we show that any rank-1 CP decomposition for a $D$-way ($D \ge 2$) tensor is unique up to  scaling indeterminacy, which implies the identifiability of the individualized layer given the population bases.
The above  condition is relatively weak compared to those in \cite{Kruskal:1989},  \cite{Sidiropoulos:2000}, and \cite {KoldaandBader:2009},  and it is easy to satisfy as it applies for an arbitrary $n$.  For example, if $n=2$ and   the factor matrices are of full rank, then the condition in Proposition \ref{prop:idt2} holds as long as $D \ge 2$.

Next we establish the theoretical properties for the recovered tensor signal $\hat{\bm{\Theta}}$ based on the observed sample tensor covariates   $\bm{\mathcal{X}}^{(m)}_i$'s.  We denote
$\bm{\gamma}=\left(\mbox{Vec}(\bm{W})', \{\mbox{Vec}(\bm{B}^{(m),d})\}'_{m,d}, \{ \bm{s}_i^{d}\}'_{i,d}  \right)'$ as the vector of all latent variable parameters. It is straightforward that $\dim(\bm{\gamma})= MR(\sum_{d=1}^Dp_d)+N(R+\sum_{d=1}^Dp_d)$.  Let  $\mathcal{I}=\{\bm{\mathcal{X}}_i^{(m)}\}_{i, m} $ denote a collection of all observed $D$-way  tensors and $|\mathcal{I}|$ be the cardinality measure of $\mathcal{I}$. That is, $|\mathcal{I}|$ denotes the number of all singlemodality images, which equals $NM$ if there is no missing.

Furthermore, let $\bm{\Theta}^{(m)}=(\theta^{(m)}_{i,j_1\cdots j_D})$ and we have  $\theta^{(m)}_{i,j_1\cdots j_D}=\sum\limits_{r=1}^{R}w_{ir}b_{j_1r}^{(m),1}\cdots b_{j_Dr}^{(m),D}+ s_{i,j_1}^{1}\cdots s_{i,j_D}^{D}$ according to the proposed  multilayer model in (\ref{eq:MIMM}).   We assume that $\mathbf{E}[X^{(m)}_{i,j_1\ldots j_D}]=\theta^{(m)}_{i,j_1\ldots j_D}$, where $X^{(m)}_{i,j_1\ldots j_D}$ denotes an element of the sample image tensor $\bm{\mathcal{X}}_i^{(m)}$, for example, a pixel in the image, and $x^{(m)}_{i,j_1\ldots j_D}$ denotes an observed value.  Then, the number of all sample elements' observations is $|\mathcal{I}|\prod_{d=1}^D p_d$, which increases as the number of images increases.   In the following, for the $(j_1,\ldots,j_D)$th element of image $\bm{\mathcal{X}}_i^{(m)}$, we define the loss function
\begin{equation*}
\setlength{\abovedisplayskip}{7pt}
\setlength{\belowdisplayskip}{10pt}
%\begin{split}
l(\mathbf{\Theta}| X^{(m)}_{i,j_1\cdots j_D}=x^{(m)}_{i,j_1\cdots j_D})=(x^{(m)}_{i,j_1\cdots j_D}-\theta^{(m)}_{i,j_1\cdots j_D})^2 =(x^{(m)}_{i,j_1\cdots j_D}-\sum_{r=1}^{R}w_{ir}b_{j_1r}^{(m),1}\cdots b_{j_Dr}^{(m),D}- s_{ij_1}^{1}\cdots s_{ij_D}^{D})^2.
%\end{split}
\end{equation*}
Consequently, we assume that  the overall objective function is an additive form of the loss function and the penalty function, that is,
\[
\setlength{\abovedisplayskip}{7pt}
\setlength{\belowdisplayskip}{10pt}
L(\mathbf{\Theta}|\bm{\mathcal{X}})=\sum_{i=1}^N \sum_{m=1}^{M} \sum_{j_1=1}^{p_1}\cdots \sum_{j_D=1}^{p_D}l(\mathbf{\Theta}| x^{(m)}_{i,j_1\cdots j_D}) +\lambda_{|\mathcal{I}|} p(\bm{\Theta}),
\]
where $\lambda_{|\mathcal{I}|}$ is a penalization coefficient.
Suppose that $\mathcal{S}$ is the parameter space of $\mathbf{\Theta}$, and that
\begin{equation} \label{minimizer}
\setlength{\abovedisplayskip}{5pt}
\setlength{\belowdisplayskip}{7pt}
\hat{\mathbf{\Theta}}=\arg\min_{\mathbf{\Theta} \in \mathcal{S}} L(\mathbf{\Theta}).
\end{equation}

In practice,  each pixel of a tensor image can only range from white to black and is usually normalized. Hence, it is sensible to assume that $\|\bm{\Theta}\|_{\infty} \le C_0 $,  $\|\bm{\gamma}\|_{\infty} \le C_1 $ and $\|\bm{\mathcal{X}}\|_{\infty} \le C_2 $ for  large constants $C_0 \ge 0$, $C_1 \ge 0$ and $C_2 \ge 0$.
Then we define the vector parameter space
$
\mathcal{S}_{\bm{\Theta}}=\{\bm{\Theta}: \|\bm{\Theta}\|_{\infty} \le C_0\}
$ and
$
\mathcal{S}_{\bm{\gamma}}=\{\bm{\gamma}: \|\bm{\gamma}\|_{\infty} \le C_1\}.
$

For each $X^{(m)}_{i,j_1\cdots j_D}$, let $l_{\Delta}(\mathbf{\Theta}|X^{(m)}_{i,j_1\cdots j_D})=l(\mathbf{\Theta},X^{(m)}_{i,j_1\cdots j_D})-l(\mathbf{\Theta}_0,X^{(m)}_{i,j_1\cdots j_D})$ be the loss difference, where $\mathbf{\Theta}_0$ corresponds to the unique true parameter. We first define:
\[
\setlength{\abovedisplayskip}{10pt}
\setlength{\belowdisplayskip}{10pt}
K(\mathbf{\Theta},\mathbf{\Theta}_0)=\frac{1}{NMp_1\cdots p_D} \sum_{i=1}^N \sum_{m=1}^{M} \sum_{j_1=1}^{p_1}\cdots \sum_{j_D=1}^{p_D} \mathbf{E}\{l_{\Delta}(\mathbf{\Theta}|X^{(m)}_{i,j_1\cdots j_D})\},
\]
which is the expected loss difference. Since $\mathbf{\Theta}_0$ is the unique true parameter, we have $K(\mathbf{\Theta},\mathbf{\Theta}_0) \ge 0$ for all $\mathbf{\Theta} \in \mathcal{S}_{\bm{\Theta}}$ and $K=0$ if and only if $\mathbf{\Theta}=\mathbf{\Theta}_0$. Therefore, we define the distance between $\mathbf{\Theta}$ and $\mathbf{\Theta}_0$ as $\rho(\mathbf{\Theta},\mathbf{\Theta}_0)=K^{1/2}(\mathbf{\Theta},\mathbf{\Theta}_0)$, and also define the variance of the loss difference as follows:
\[
\setlength{\abovedisplayskip}{10pt}
\setlength{\belowdisplayskip}{10pt}
V(\mathbf{\Theta},\mathbf{\Theta}_0)=\frac{1}{NMp_1\cdots p_D} \sum_{i=1}^N \sum_{m=1}^{M} \sum_{j_1=1}^{p_1}\cdots \sum_{j_D=1}^{p_D} \mbox{Var}\{l_{\Delta}(\mathbf{\Theta}|X^m_{i,j_1\cdots j_D})\}.
\]

Under the $L_2$-loss, it is expected that $K(\mathbf{\Theta},\mathbf{\Theta}_0)=\frac{1}{NMp_1\cdots p_D}\|\mathbf{\Theta}-\mathbf{\Theta}_0\|_F^2$, and that $V(\mathbf{\Theta},\mathbf{\Theta}_0)=\frac{4\sigma^2}{NMp_1\cdots p_D}\|\mathbf{\Theta}-\mathbf{\Theta}_0\|_F^2$, where $\sigma^2$ is assumed to be the same variance of each element of the tensor. %Let

\begin{thm} \label{thm:tensor}
Suppose $\hat{\mathbf{\Theta}}$ is the sample estimator satisfying (\ref{minimizer}),  then we have
\[
\setlength{\abovedisplayskip}{7pt}
\setlength{\belowdisplayskip}{10pt}
\mathbf{P}(\frac{1}{|\mathcal{I}|^{1/2}}\|\hat{\mathbf \Theta}-\mathbf \Theta_0\|_F \ge \tau_{|\mathcal{I}|}) \leq 7\exp(-c_1|\mathcal{I}| \tau_{|\mathcal{I}|}^2)
\]
for $\tau_{|\mathcal{I}|} = \max (\varepsilon_{|\mathcal{I}|},\lambda^{1/2}_{|\mathcal{I}|})$,   where $c_1 > 0$  is a constant and $\varepsilon_{|\mathcal{I}|} \sim {|\mathcal{I}|}^{-1/2}$. The best possible rate  is achieved at $\tau_{|\mathcal{I}|}=|\mathcal{I}|^{-1/2}$ when $\lambda_{|\mathcal{I}|} \sim \varepsilon^2_{|\mathcal{I}|}$.
\end{thm}

Theorem \ref{thm:tensor} indicates that, with an appropriate rate of the penalty term shrinking to zero, the recovered tensor signal in (\ref{minimizer}) converges to the true one as the number of images goes to infinity. In other words, assuming the identifiability conditions are satisfied, Theorem \ref{thm:tensor} ensures that the extracted information from the sample images captures the true underlying information as the sample size increases. In addition, the  convergence rate is established under the $L_2$ distance, which is  a special case of the Kullback-Leibler divergence.

\noindent \textbf{Remark 1}:
In practice, the estimator specified  in (\ref{minimizer}) is a global minimizer and may not be attainable due to the non-convex nature of the proposed optimization problem. One possible solution is to follow (1.1) of \cite{Shen:1998} and relax (\ref{minimizer}) to allow an approximate global minimizer,  in the sense that the  IMTL estimator approaches the global minimizer   asymptotically, which is less restrictive on the requirement of the global optimum.   Another solution is to develop theoretical properties directly for the IMTL estimator, which is viable because of the block-wise convexity of the proposed criterion function \citep{Li:2017, Zhu:2017}. However, this  may require  good initial values in order to establish  the consistency property.

Next, we investigate the theoretical property of the  predictive model with the extracted sample information.
In the proposed framework, we consider a generalized linear model \citep{McCullagh:1983} assuming that the response variable  $Y_i$ is associated to the true image signal ($\bm{\Theta}_i$) via
$
\mathbf{E}(Y_i|\bm{\Theta_i})=\mu(\eta_i) \quad \text{and}\quad \eta_i=f(\bm{\Theta_i})^T \bm{\beta},
$
where $f: \mathbb{R}^{p_1\times  \ldots \times p_D \times M} \rightarrow \mathbb{R}^{p_f}$ is the feature-extraction mapping following (\ref{eq:MIMM}), and $\mu(\cdot)$ is a link function.  Let $F(\bm{\Theta}_{|\mathcal{I}|})= (f(\bm{\Theta}_{1})^T, \ldots,f(\bm{\Theta}_{N})^T)^T$ denote the extracted feature covariates based on the true imaging signals, and  let  $\mathcal{L}\left(y_i, f(\bm{\Theta}_{i}), \bm{\beta}\right)= \mathcal{L}(y_i, \eta_i\{f(\bm{\Theta}_i), \bm{\beta}\})$ denote a general loss function analogues to that   in (\ref{eq:glm}).  Consequently, a  general criterion function  for supervised learning  with  true imaging information can be presented as
\begin{equation}\label{eq:super}
\setlength{\abovedisplayskip}{5pt}
\setlength{\belowdisplayskip}{7pt}
G_{N}(\bm{y}, F(\bm{\Theta}_{|\mathcal{I}|}), \bm{\beta})= \frac{1}{N}\sum_{i=1}^N \mathcal{L}(y_i, f(\bm{\Theta}_{i}), \bm{\beta})+ \lambda_{\bm{\beta}} \cdot p^*(\bm{\beta}),
\end{equation}
where $p^*(\bm{\beta})$ is a penalty term imposed on feature covariates effect $\bm{\beta}$, $\lambda_{\bm{\beta}}$ is the associated tuning parameter,  and $\bm{y}=(y_1,\ldots, y_N)'$ is the sample response.

Note that the model in (\ref{eq:super}) essentially refers to a ``true'' model with the underlying true imaging information. Since  only sample images with noises are  observed,  we actually  use  the recovered  information $\hat{\bm{\Theta}}_{|\mathcal{I}|}$ and the corresponding  sample features $F(\bm{\mathcal{X}}_{|\mathcal{I}|})=F(\hat{\bm{\Theta}}_{|\mathcal{I}|})$ in  the supervised learning model (\ref{eq:super}), leading  to a sample objective function.
Define $\Omega_{\bm{\beta}}$  and $\Omega_{\eta}$ as the parameter spaces for $\bm{\beta}$ and $\eta_i$ ($1\le i \le N$), respectively.
In the following, we introduce a regularity condition on the loss function $\mathcal{L}(\cdot)$:

\noindent (\textbf{C1}): For $\eta_i,\tilde{\eta}_i \in \Omega_{\eta}$ and all $y_i$'s,  there exists a function $K_i(\cdot)$ such that
\[
\setlength{\abovedisplayskip}{7pt}
\setlength{\belowdisplayskip}{7pt}
 | \mathcal{L}(y_i, \eta_i) -  \mathcal{L}(y_i, \tilde{\eta}_i)| \le K_i(y_i) | \eta_i - \tilde{\eta}_i |, \quad 1 \le i \le N,
\]
and $\frac{1}{N}\sum_{1}^N \mathbf{E} [K_i(y_i)^2] \le K_0$ holds for a positive constant $K_0$.

\begin{lemma}\label{lm:dfloss}
Suppose condition (C1) holds. Under regularity condition (R1) provided in supplementary material  (A.5), and if $\frac{1}{N^{1/2}}\| F(\bm{\mathcal{X}}_{|\mathcal{I}|}) - F(\bm{\Theta}_{|\mathcal{I}|}) \|_F \rightarrow 0$,  then, for any given $\bm{\beta} \in \Omega_{\bm{\beta}}$, we have
\[
\setlength{\abovedisplayskip}{7pt}
\setlength{\belowdisplayskip}{10pt}
 \bigg| G_{N}(\bm{y}, F(\bm{\mathcal{X}}_{|\mathcal{I}|}), \bm{\beta}) - G_{N}(\bm{y}, F(\bm{\Theta}_{|\mathcal{I}|}), \bm{\beta}) \bigg| \rightarrow_p 0.
\]
\end{lemma}

Lemma \ref{lm:dfloss} indicates that the sample prediction model approaches the  true model asymptotically as long as the recovered imaging signal converges to the true one.
Next, we investigate the asymptotic  property of the sample model estimation.  Let $\bm{\beta}_0$ be the true value of $\bm{\beta}$  denoting the effect of the true imaging signal,  let  $\hat{\bm{\beta}}(\bm{\mathcal{X}}_{|\mathcal{I}|}) = \arg\min\limits_{\bm{\beta}} G_N(\bm{\beta}| \bm{y}, F(\bm{\mathcal{X}}_{|\mathcal{I}|}))$ be the sample estimator with observed image covariates, and let  $\tilde{\bm{\beta}}( \bm{\Theta}_{|\mathcal{I}|}) {=} \arg\min\limits_{\bm{\beta}} G_N(\bm{\beta} | \bm{y}, F(\bm{\Theta}_{|\mathcal{I}|}))$ be the oracle estimator with true imaging signals.
Furthermore,  we denote  $\mathcal{R}(\hat{\mathcal{L}}_N) = \frac{1}{N}\sum_{i=1}^N \mathcal{L}(y_i, f(\bm{\mathcal{X}}_{i}), \hat{\bm{\beta}}(\bm{\mathcal{X}}_{|\mathcal{I}|}))$ as the empirical risk for the sample loss function, and denote $\mathcal{R}(\tilde{\mathcal{L}}_N) = \frac{1}{N}\sum_{i=1}^N \mathcal{L}(y_i, f(\bm{\Theta}_{i}), \tilde{\bm{\beta}}(\bm{\Theta}_{|\mathcal{I}|}))$ as the empirical risk for the oracle loss function with  true imaging information.  The next results provide the estimation consistency of the  imaging covariates' effect as well as the  empirical error.

\begin{thm}\label{thm:pred}
Suppose the conditions in Theorem \ref{thm:tensor} and the identifiability conditions in Proposition \ref{prop:idt2} hold. Under regularity conditions (R1-R4) (supplementary material  A.5) and  condition (C1), if $\lambda_{\bm{\beta}} \rightarrow 0$,  as $|\mathcal{I}|\rightarrow \infty$, we have
$$
\setlength{\abovedisplayskip}{3pt}
\setlength{\belowdisplayskip}{3pt}
|\hat{\bm{\beta}}(\bm{\mathcal{X}}_{|\mathcal{I}|})-
\tilde{\bm{\beta}}(\bm{\Theta}_{|\mathcal{I}|})| \rightarrow_p 0 \quad  \text{and} \quad \hat{\bm{\beta}}(\bm{\mathcal{X}}_{|\mathcal{I}|}) \rightarrow_p \bm{\beta}_0,
$$
and also,
$$
\setlength{\abovedisplayskip}{3pt}
\setlength{\belowdisplayskip}{3pt}
\big|\mathcal{R}(\hat{\mathcal{L}}_N)- \mathcal{R}(\tilde{\mathcal{L}}_N)\big| \rightarrow_p 0.
$$
\end{thm}

In addition to  model estimation consistency,  Theorem \ref{thm:pred} indicates that  the empirical error of the sample predictive model approaches the oracle one, where the true imaging signals are known.
With an appropriately selected  predictive model in (\ref{eq:glm}),  the proposed approach is able to achieve a consistent and efficient prediction result.

\noindent \textbf{Remark 2}:  The exact generalization error bound of the proposed model actually depends on a particular predictive model specified in (\ref{eq:super}) as well as the response types.  Consider a new observation with true signal  $(Y^*_j, \bm{\Theta}^*_j)$, in a generalized linear model, Theorem 2 indicates that we would have a consistent mean response prediction  $\hat{\mu}^*_j = \mu(f(\bm{\Theta}^*_j)^T \hat{\bm{\beta}}(\bm{\mathcal{X}}_{|\mathcal{I}|})) \rightarrow_p \mathbf{E}[Y_j^*|\bm{\Theta}_j^*] $, where $\mu(\cdot)$ is a  link function.  Furthermore,  for a  continuous and bounded  response, the $L_2$-prediction-risk $\mathbf{E}(\hat{Y}_j^*{-}Y_j^*)^2$ converges to the theoretical lower bound $\mbox{Var}_{\bm{\Theta}_j^*}(Y^*_j)$ with $\hat{Y}^*_j=\hat{\mu}^*_j$; while  for a binary response $Y^*_j$, let $\hat{\pi}(\bm{\Theta}_j^*)=\mathbf{1}[\hat{\eta}(\bm{\Theta}_j^*)>1/2]$ be the trained classification rule, where $\hat{\eta}(\bm{\Theta}_j^*)=\mbox{logit}(f(\bm{\Theta}^*_j)^T \hat{\bm{\beta}})$,  then we have $\hat{\pi}(\bm{\Theta}_j^*)\rightarrow_p \pi^0_j$,  the optimal Bayes' rule.

\section{Simulation  Studies}\label{sec:simB}

In this section, we  investigate the performance of the proposed approach comparing with other competing methods through a simulation study of heterogeneous weak signals, which mimics the real data structure and also frequently arise in clinical diagnosis and cancer imaging analysis.   Specifically, we simulate  four-modality imaging data,  where multiple modalities imaged  from the same individual share  random-location heterogeneous signals,  while each modality  contains its unique background bases.  Additional simulations regarding various heterogeneous signal patterns with the singlemodality data and the impact of the contrast between population-shared components and individualized  components are provided in the supplementary materials Section B.

In this study,  we simulate  the $m$th-modal image for the $i$th subject as   $\bm{\mathcal{X}}^{(m)}_i=\bm{\mathcal{A}}^{(m)}_{i}+\bm{\mathcal{B}}_i+\bm{\mathcal{N}}_i$, $m=1,\ldots,4$, consisting of  the true signal feature image $\bm{\mathcal{B}}_i$, the modality-specific background image $\bm{\mathcal{A}}^{(m)}_i$ and the noise background image $\bm{\mathcal{N}}_i$.
For each subject, we randomly select $s_i$ pixels  of $\bm{\mathcal{B}}_i$ to be valued of  $2$ (signal pixels) with the other pixels of 0.   We generate the response label $y_i$ from a Bernoulli distribution with a probability of $0.4$ to be $1$.   The number of the signal pixels $s_i$ is generated from a Poisson distribution with means $\mu_C=25$ and $\mu_N=5$ for the cancer subject's image (given $y_i=1$) and the normal subject's image (given $y_i=0$), respectively. The noise elements of  $\bm{\mathcal{N}}_i$ are generated from $N(0,0.2^2)$.

\begin{figure}[H]
   \centering
   \includegraphics[width=1\linewidth]{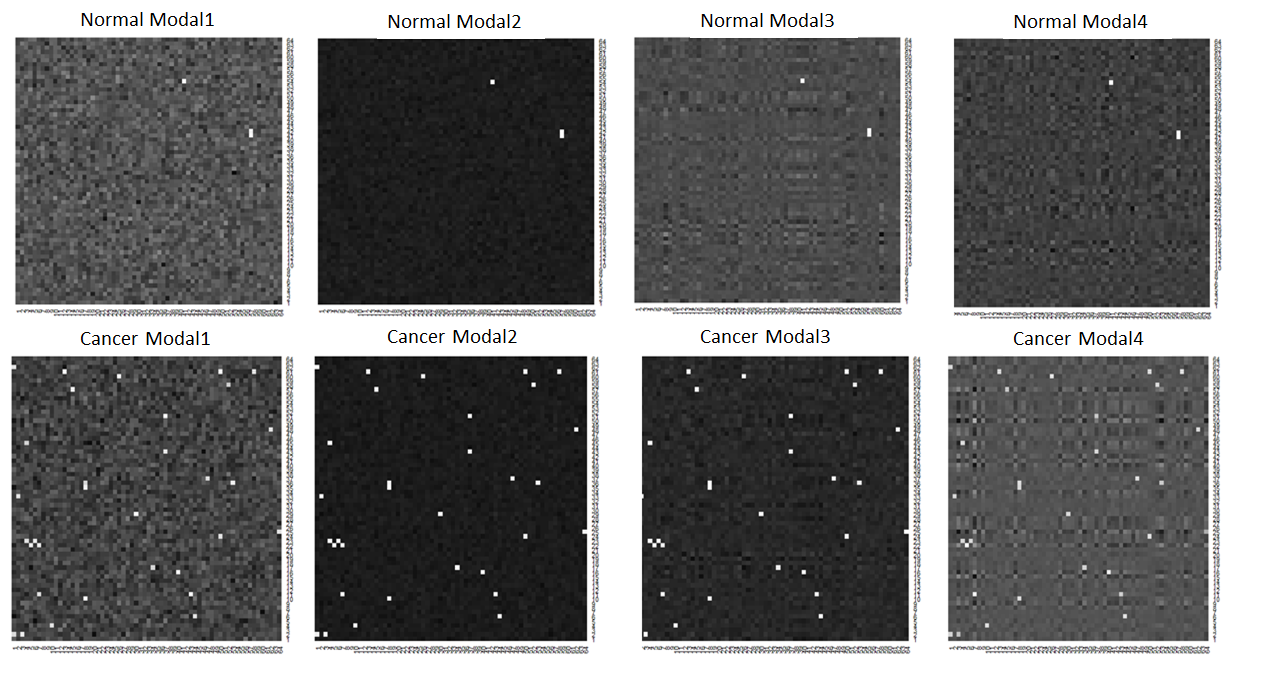}
   \caption{The representative four-modality images for a normal subject and a cancerous subject in simulation study, with each modal image size of $64\times64$.}
   \label{fig:simMM}
\end{figure}

Moreover, the first modality background $\bm{\mathcal{A}}^{(1)}_{i}$ is  a random noise matrix with elements generated from  $N(0,0.1^2)$; the second modality has a  uniform background with  $\bm{\mathcal{A}}^{(2)}_{i}=w^{(2)}_{i}\mathbf{1}_D\mathbf{1}_D^T$, where $\mathbf{1}_D$ is a $D \times 1$ vector of 1's and $w^{(2)}_{i}$ is generated from $N(0,0.1^2)$; and both the third and fourth modality imaging  have low-rank structures with $\bm{\mathcal{A}}^{(m)}_{i}=\sum_{r=1}^5 w^{(m)}_{ir} \bm{a}^{(m),1}_{r}\circ\bm{a}^{(m),2}_{r}$ ($m=3,4$), where $w^{(3)}_{ir}$ and  $w^{(4)}_{ir}$ are generated from $N(0,0.1^2)$, $\bm{a}^{(m),1}_{r}$'s and $\bm{a}^{(m),2}_{r}$'s are generated from $N(\bm{0}, \mathbf{I}_D)$ and $N(\bm{0}, 0.5\mathbf{I}_D)$, respectively,  and $\mathbf{I}_D$ is the $D$-dimensional identity matrix.  We set the sample size as $60$ and $100$, equally for the training set, the validation set and the testing set, and the marginal imaging  dimension as $D=64$.    Figure \ref{fig:simMM} illustrates sample normal and cancerous images.

To evaluate the prediction performance, we calculate the  prediction accuracy rate ($\mathbf{P}[\hat{y}=y]$), the sensitivity ($\mathbf{P}[\hat{y}=1|y=1]$) and the specificity  ($\mathbf{P}[\hat{y}=0|y=0]$) on the testing test.   We compare the proposed IMTL model with the higher-order CP decomposition method (HOCPD) in (\ref{eq:hocpdb}), the marginal principal component analysis (MPCA) method by mimicking \cite{Caffo:2010},  the vectorizing  $L_1$-penalized logistic regression model (VPL), the tensor regression model \citep[TR,][]{Zhou:2013} and the convolutional neural network (CNN).

The VPL is applied on a $16,384$-dimensional vector predictor by vectorizing all four modalities  and then fits an $L_1$-penalized logistic regression model for the binary response, which is implemented by the R package \emph{``glmnet''} \citep{Friedman:2010}.
The MPCA first extracts feature from each individual modality separately and then fits a logistic model with all modal features. The TR and the HOCPD are applied on the integrated multimodality image (third-order tensor predictor).
The TR method  is implemented by \cite{ZhouM:2013}'s Matlab toolbox \emph{``TensorReg''}.   In the following numerical studies and real data analysis, we  employ an $L_1$-penalized logistic regression model for the IMTL, the HOCPD and the MPCA methods at the predictive stage using the extracted features.  The relevant tuning parameters associated with  each model are selected through minimizing  the prediction error rates ($\mathbf{P}[\hat{y}\ne y]$) on the validation set, respectively.

In particular, the CNN method is implemented by Matlab toolbox \emph{``matcovnet''} \citep{Vedaldi:2015}, which inputs four modalities as four channels.    Since the tuning of the CNN is crucial,   we tune the CNN  to minimize the classification error rates on the validation set  over the number of layers, the number of filters on each convolution layer, the filter size, depth and  the stride size,  the pooling window size, the pooling stride size, the pooling method, the activation function, the number of epochs and the learning rates.  The detailed CNN architectures and the tuning process   in  our numerical studies  are provided in the supplementary material Section C.  In addition, we also implement  the TensorFlow for the CNN  and the results are consistent with the Matlab's results.

Table \ref{tb:simMM} provides the prediction results based on 100 replications.  The proposed method (IMTL) outperforms  other methods with the highest prediction accuracy ($0.86$ and $0.94$) and sensitivity ($0.69$ and $0.88$) for both sample sizes ($60$ and $100$), respectively.  Moreover,  the  IMTL,  the HOPCD methods and  the CNN,  have  significant  advantages over the VPL and the MPCA methods which assume the independence between the four modalities.  This indicates that integrating different modalities' information at feature extraction  enhances the prediction power. Furthermore, the proposed IMTL method achieves more than $16\%$ improvement in prediction accuracy than the HOCPD,  indicating that the proposed method is more effective in utilizing correlation information among different modalities with additional individualized layers.  An increased training data size improves the CNN's prediction power, however, the IMTL still outperforms the CNN with a $10\%$ higher overall accuracy and a $13.4\%$ higher sensitivity at a sample size of 100.

Although the CNN is a  powerful tool for image classification,  it requires a large number of training samples to entails multiple hidden layers and involves a large number of parameters \citep{Keshari:2018, Wagner:2013, Yu:2017b, Yu:2017a}.  In practice, the sufficient sample size is not always attained, especially for cancer imaging data.   In addition, the heterogeneity is another great challenge occurred in   cancer imaging, while the CNN is less capable of capturing heterogeneous signals with relatively weak strength.

\begin{figure}[H]
   \centering
   \includegraphics[width=1\linewidth, height=0.5\linewidth]{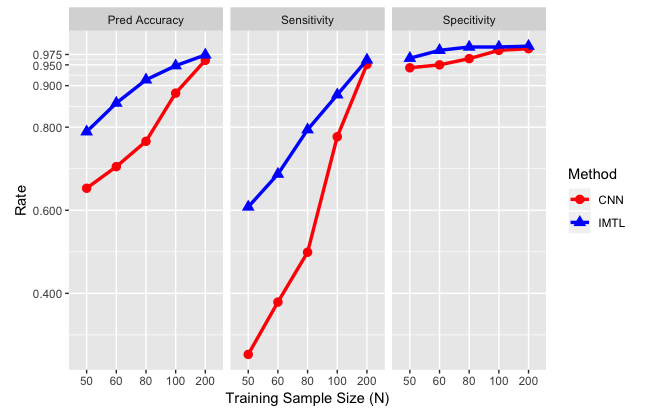}
   \caption{Prediction performance of the CNN and the IMTL  under the settings of simulation study (Section 5) with different  sample sizes. The results are based on 50 replications.  }
   \label{fig:sizeC}
\end{figure}

We conduct additional simulations to  investigate the performance of the CNN compared with the proposed IMTL in various situations.  First,  we compare the performance of the CNN and the IMTL,  with different training sample sizes. Figure  \ref{fig:sizeC} shows that the IMTL consistently outperforms the CNN, especially when the training sample size is limited, while the CNN gradually ``catches up'' as the sample size increases.  This confirms that the CNN  requires a large  size of training samples in order to perform well.  The image size in this  simulation is  only   $64 \times 64$.  According to the trend  in Figure \ref{fig:sizeC},  for real data with an image size of hundreds to thousands, the CNN could  require quite a substantial  number of training samples to guarantee a good  performance.

\begin{figure}[H]
   \centering
   \includegraphics[width=1\linewidth, height=0.5\linewidth]{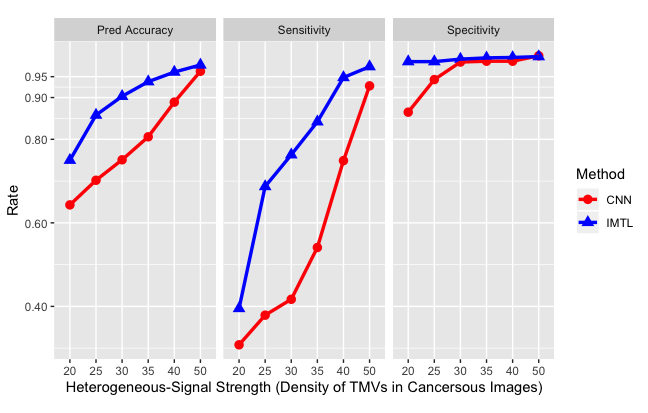}
   \caption{Prediction performance of the CNN and the IMTL  under the settings of simulation study (Section 5) with different  heterogeneous-signal strengths. The signal strength and the heterogeneity level are controlled by the density (mean number: $\mu_C$) of the TMVs in a cancerous  image, while the density in a normal image ($\mu_N$) is fixed as 5. The training sample size is set as 60 and the results are based on 50 replications.}
   \label{fig:signalC}
\end{figure}

Second,  we  investigate the performance of the CNN and the proposed IMTL with varying  heterogeneous-signal strengths.   To  mimic   real data, we utilize the concentration intensity of the TMVs to measure the strength and the heterogeneity of  signals, which is controlled by the mean value of the number of the TMVs ($\mu_C$) in cancer  images,  while fixing the normal images' mean ($\mu_N=5$).  Therefore, the larger the $\mu_C$ is, the stronger the signal contrast  between the cancerous images and the normal images is.  On the other hand,  if  $\mu_C$ is larger, then the heterogeneity is  less since  the TMVs tend to  fill the entire cancer image, and  thus the heterogeneity of the  randomly-located signals is reduced.  We set the training sample size as 60. Figure \ref{fig:signalC} shows the prediction power of the CNN and the IMTL with a varying $\mu_C$. It is clear that the IMTL outperforms the CNN consistently, while the  CNN improves as the signal strength becomes stronger and the heterogeneity reduces.  This  simulation confirms that  the proposed IMTL is more powerful than the CNN  in  situations where  the heterogeneity is high and the signal strength is weak.

\section{Real Data: Multiphoton Imaging Data for Breast Cancer}
We apply  the proposed method to the multimodality optical imaging data for breast cancer produced by Boppart Lab \citep{Tu:2016} at University of Illinois  Urbana-Champaign.
To better visualize the tissue's biological structure at cellular and molecular levels,  \cite{Tu:2016}'s multiphoton microscope generates multimodal images through emission of different numbers of photons. Specifically, there are two-photon auto-fluorescence (2PAF), three-photon auto-fluorescence (3PAF), second-harmonic generation (SHG) and third-harmonic generation (THG).   Two-photon-fluorescence microscopy is commonly used to visualize tissue morphology and physiology at a cellular level, and three-photon-fluorescence with  longer wavelength  can reach deeper levels of the tissue and  thus provide higher imaging resolution  \citep{Chu:2005, Horton:2013}. This  new technique is able to capture the important tumor-associated microvesicles (TMVs) which are not easily  identified by conventional imaging tools such as  histology imaging.

\begin{figure}[H]
     \centering
     \includegraphics[width=1\textwidth]{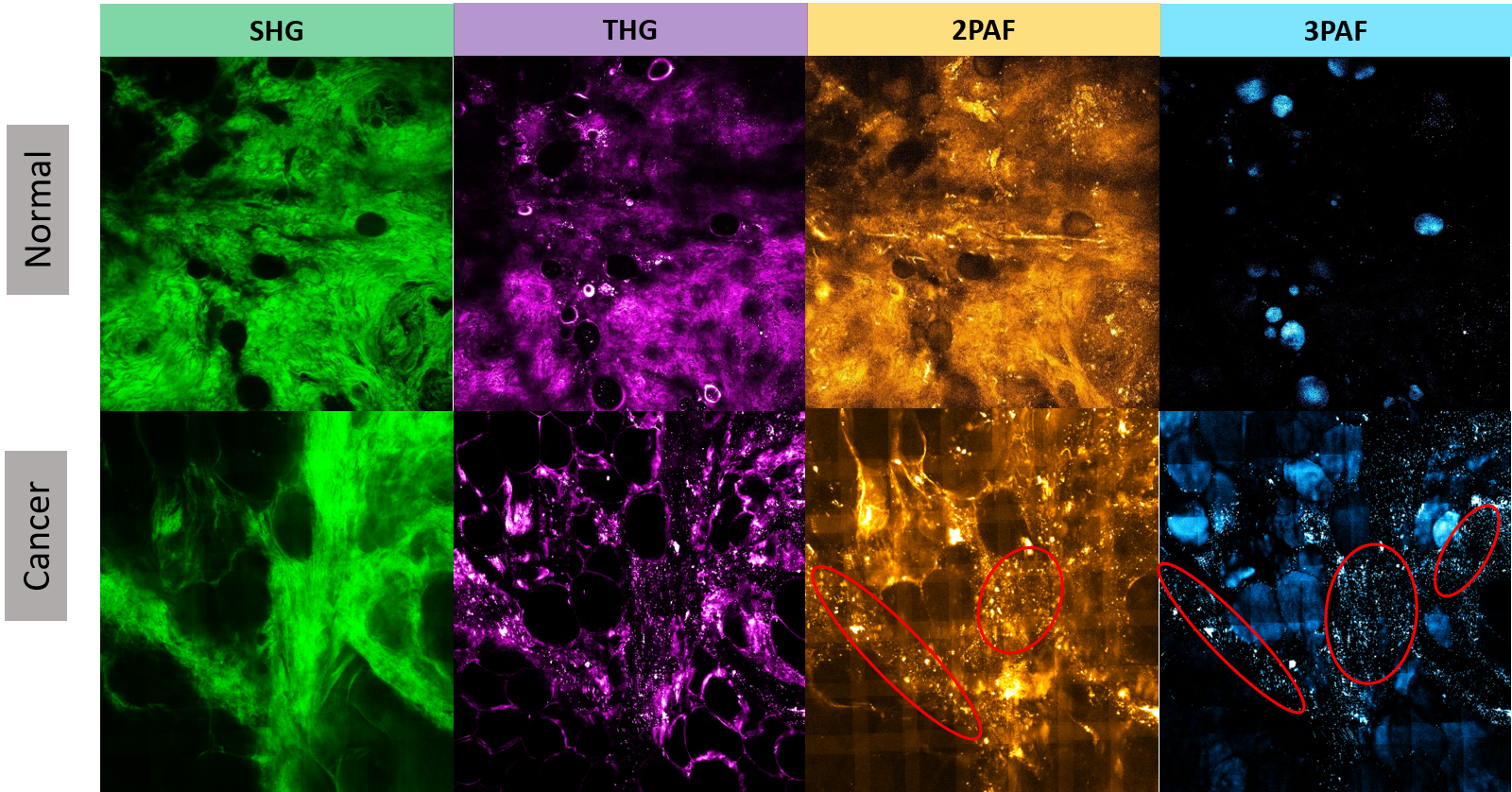}
     \caption{The four-modality microscope images for a normal human subject ($2382 \times 2401$) and a cancerous human subject ($2191 \times 2193$). See circles of TMVs.}
     \label{fig:human}
\end{figure}

Figure \ref{fig:human}  illustrates the four modalities  for a human subject's normal breast tissue and a  human subject's cancerous breast tissue.  In contrast to the normal tissue,  multiple modalities clearly indicate a large number of TMVs which  spread out in the microenvironment on the cancerous imaging,  particularly in the 2PAF image and the 3PAF image.  The normal imaging also presents microvesicles in the 3PAF image,  but they are more sporadic with much less intensity.

Prior knowledge in cancer detection shows that informative TMVs are frequently observed in the microenvironment between certain biological organizations such as at the lipid boundary area and around the stroma \citep{Tu:2016}. Therefore, we study  segmented imaging from three normal human individuals and two cancerous individuals. Specifically, we generate sample images through segmentation,  which leads to $107$ normal subjects and $53$ cancerous subjects, each subject having four modalities of images, with each modality of $100\times100$ pixels.
To fit the predictive models and evaluate the diagnosis performance,  we randomly split the total sample into a training set, a validation set and a testing set with $60$, $40$ and $60$ subjects, respectively.  The prediction results are evaluated by  prediction accuracy, sensitivity and specificity  on the testing set  based on $30$ replications.

\begin{figure}[H]
   \centering
   \includegraphics[width=1\linewidth, height=0.5\linewidth]{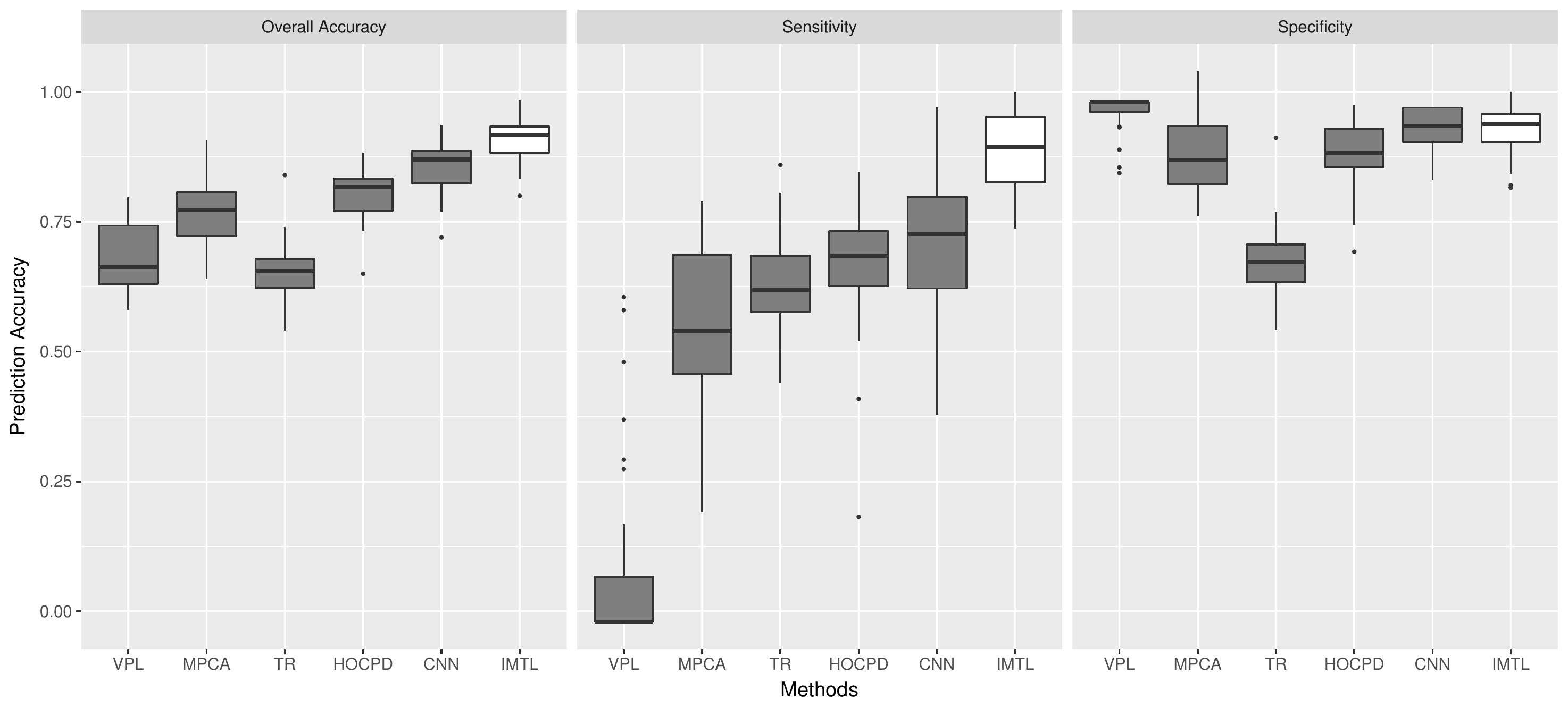}
   \caption{The boxplots of the prediction results for the VPL, the TR, the MPCA, the HOCPD, the CNN and the IMTL models utilizing all four modalities, based on 30 random splitting replications. Sample sizes for the training set, the  validation set and the testing set are 60, 40 and 60, respectively.}
   \label{fig:realbox1}
\end{figure}

We compare the proposed IMTL method to the five methods described in Section \ref{sec:simB}.   Table \ref{tb:realdata} provides the averaged prediction results on the testing set,  which indicates that the proposed   IMTL method outperforms the other methods  significantly in terms of  achieving  the  highest overall prediction accuracy  and sensitivity. Specifically, the prediction accuracy of the proposed method improves  by 34\%, 39\%, 19\%, 13\% and 7\% compared to the VPL, TR, MPCA, HOCPD and CNN approaches, respectively. The boxplot in Figure \ref{fig:realbox1} suggests that the proposed method also has the smallest standard error in overall prediction accuracy.  Note that the outstanding predictive power of the proposed method is mainly due to its highest sensitivity score, which is also of  primary interest for  cancer diagnosis. In practice, the proportion of the potential cancer patients is rather small compared to the general population, and  correctly  detecting cancer at earlier stages of cancer development is crucial and critical.

\begin{figure}[H]
   \centering
   \includegraphics[width=1\linewidth,height=0.5\linewidth]{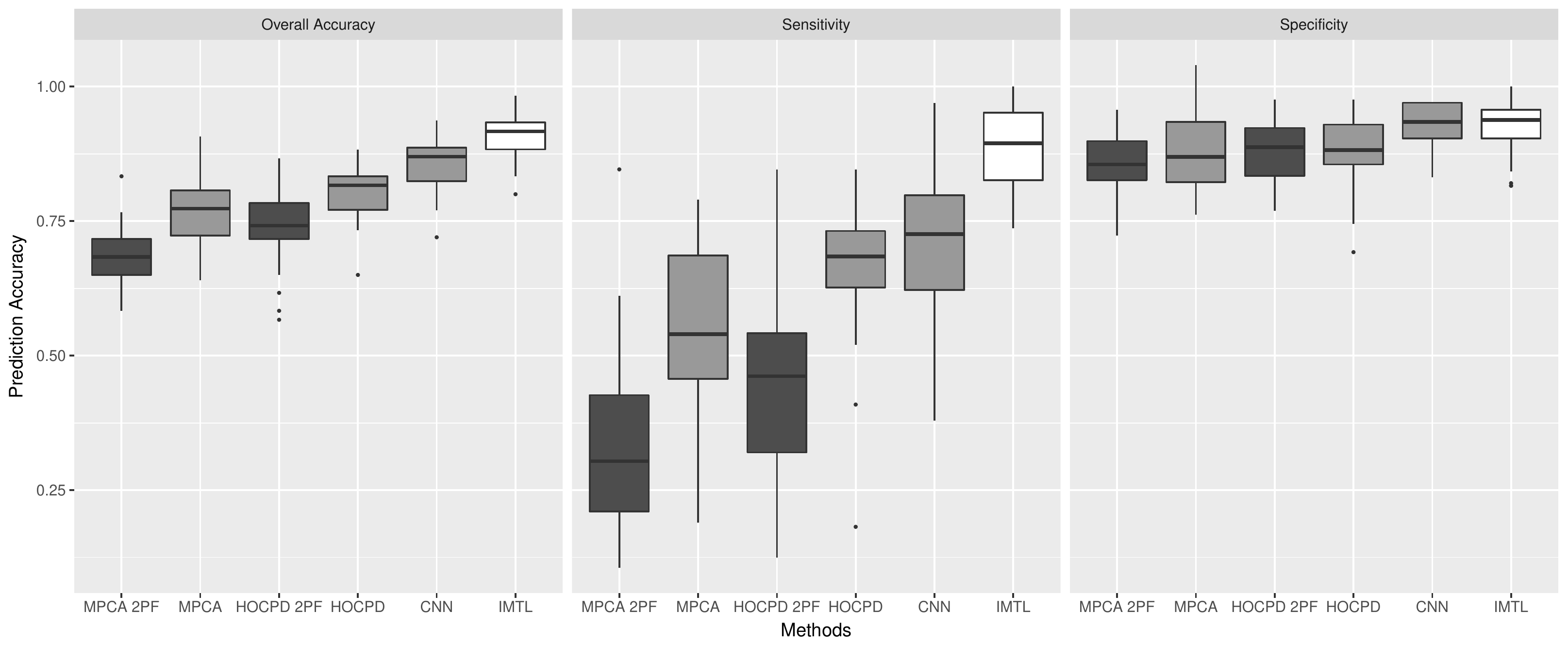}
   \caption{The boxplots of the prediction results for the MPCA model and the HOCPD model using only 2PAF modality (MPCA 2PF and HOCPD 2PF, respectively), the MPCA and the HOCPD model using all modalities,  the CNN and the IMTL models using all modalities. The results are based on 30 random splitting replications. Sample sizes for the training set, the  validation set and the testing set are 60, 40 and 60, respectively. }
   \label{fig:realbox2}
\end{figure}

For this data, the VPL method and the tensor regression  model  perform inadequately compared to other methods due to the random location and the weak signal strength of the TMVs.  Moreover, without incorporating   individual-wise heterogeneity, the MPCA and the HOCPD are  inefficient  with low sensitivity.  In general, the CNN provides an acceptable prediction result,  however, due to limited sample size and heterogeneous imaging features,  the CNN is not as effective  in identifying cancerous subjects with only a 73\% sensitivity rate compared to the IMTL's 90\% sensitivity rate.  In addition, Figure \ref{fig:realbox2} provides an illustration of the prediction performance using a single modality only compared to utilizing all modalities.  Figure \ref{fig:realbox2} clearly shows that integrating multiple modalities improves the predictive power, especially in detecting cancerous subjects.

\section{Discussion}
In this article, we propose an individualized multilayer tensor learning model  incorporating multimodality imaging tensor covariates to predict targeted responses.  In the proposed  model, we extract both individual-specific and population-wise important features simultaneously from higher-order tensor covariates through different layers, and then fit a prediction model with the extracted features.  We illustrate the proposed method through numerical studies and human breast cancer imaging application  on both singlemodality and multimodality data.

A major contribution of the proposed method is that we achieve heterogeneous tensor decomposition through utilizing an individualized layer in addition to population-shared modality-specific structure.  Our method is motivated by a  multimodality imaging study for breast cancer diagnosis, where the biomarker TMVs' are distributed with great heterogeneity and each modality has its unique background features.
Most existing methods assuming homogeneous structure on signals' features are  either  infeasible or inefficient in our situation.  In contrast, the proposed different layers are capable of capturing   individual-specific  spatial features  through  integrating different modalities'  imaging information for the same individual.  Both numerical studies and theoretical results  demonstrate that the  proposed method can achieve higher diagnostic accuracy.

 In the proposed method,   multilayer tensor decomposition  in the first stage is not connected to the response variable directly, which may not guarantee an optimal feature extraction for  supervised learning in the second stage.  For future research, it is worth  developing   a supervised   feature extraction scheme which can be more powerful in  predicting outcomes.  We point out  three potential  directions. The first one is to   optimize  both the classification loss and the  image tensor reconstruction  loss,  where the latter can be treated as a part of regularization. The second possible direction  is to incorporate sufficient dimension reduction techniques to search the  ``best'' bases layers conditional on the response variable. The third  direction  is to link the feature extraction stage and the prediction stage in a hierarchical form mimicking the CNN framework, and update the first stage of feature extraction adaptively based on the prediction loss.

\begin{flushleft}\large
 \textbf{Supplementary Materials}\\
\end{flushleft}
The online supplement contains all technical proofs, additional numerical results and computation details.

\begin{flushleft}\large
 \textbf{Acknowledgments}\\
\end{flushleft}
The authors would like to acknowledge support for this project from the National Science Foundation grants DMS-1415308, DMS-1613190, DMS-1821198. The authors are grateful to reviewers,  the Associate Editor and Editor for their insightful comments and suggestions which have  improved the manuscript significantly.

%reference
\bibliographystyle{APA}
\bibliography{IMTL_bib}
%\restoregeometry

\newpage

\begin{table}[H]
\centering
\caption{\small  Prediction results  (standard errors in subscripts) of the proposed method (IMTL) compared with five other competing methods for simulation study, based on 100 replications. Equal sample size for training set ($N_{tr}$), validation set ($N_{vl}$) and testing set ($N_{ts}$).
}
\label{tb:simMM}
\begin{tabular}{llcccccc}
\toprule
$\bm{N_{tr}}$ &             &  VPL   &TR  &MPCA   &HOCPD  &CNN& IMTL  \\\hline
\multirow{3}{*}{$60$}
&Pred Accuracy & $0.593_{0.051}$ & $0.573_{0.054}$ &$0.598_{0.078}$ &$0.731_{0.175}$ &$0.714_{0.083}$&{$\bm{0.858}_{0.069}$}\\
&Sensitivity   & $0.000_{0.000}$ & $0.415_{0.103}$ &$0.443_{0.143}$  &$0.526_{0.229}$ & $0.379_{0.209}$& {$\bm{0.687}_{0.119}$}\\
&Specificity   & $1.000_{0.000}$ & $0.674_{0.097}$ &$0.691_{0.080}$  &$0.915_{0.094}$& $0.955_{0.023}$& {$\bm{0.985}_{0.064}$}\\\hline
\multirow{3}{*}{$100$}
&Pred Accuracy & $0.608_{0.043}$ & $0.603_{0.045}$ &$0.634_{0.057}$ &$0.808_{0.168}$ &$0.884_{0.069}$&{$\bm{0.941}_{0.041}$}\\
&Sensitivity   & $0.001_{0.005}$ & $0.453_{0.090}$ &$0.490_{0.113}$  &$0.682_{0.226}$ & $0.774_{0.160}$& {$\bm{0.878}_{0.064}$}\\
&Specificity   & $0.998_{0.008}$ & $0.697_{0.091}$ &$0.728_{0.070}$  &$0.931_{0.064}$& $0.985_{0.010}$& {$\bm{0.993}_{0.030}$}\\
\bottomrule
\end{tabular}
\end{table}

\vspace*{2cm}

\begin{table}[H]
\centering
\caption{\small Prediction results of the proposed method (IMTL) compared with five other competing methods for human breast cancer imaging data,  based on 30 random  replications. Sample size for the training set, the  validation set and the testing set are 60, 40 and 60, respectively.}

\label{tb:realdata}
\begin{tabular}{lcccccc}
\toprule
Model            &  VPL   &TR  &MPCA    &HOCPD &CNN &IMTL  \\\hline
Pred Accuracy & 0.681  & 0.656 & 0.766    & 0.803  &0.871 & \textcolor[rgb]{1.00,0.00,0.00}{\textbf{0.921}} \\
Sensitivity      & 0.086& 0.628&0.535 & 0.656 &0.729& \textcolor[rgb]{1.00,0.00,0.00}{0.903} \\
Specificity      & 0.962 &0.671 &0.875 & 0.878 & 0.928 & \textcolor[rgb]{1.00,0.00,0.00}{0.925} \\
\bottomrule
\end{tabular}
\end{table}

\newpage

\end{document}